\pdfoutput=1

\documentclass[11pt]{article}
\usepackage{acl}

\usepackage{times}
\usepackage{latexsym}
\usepackage[T1]{fontenc}
\usepackage[utf8]{inputenc}
\usepackage{microtype}
\usepackage{multirow}
\usepackage{graphicx}
\usepackage{paralist}
\usepackage{covington}
\usepackage{attrib}
\usepackage{comment}
\usepackage{amsmath}
\usepackage{pdflscape}
\usepackage{tabulary}
\usepackage{comment}

\title{A Systematic Comparison of Contextualized Word Embeddings for \\ Lexical Semantic Change
\\ {\small \texttt{Submitted to NAACL 2024}}}

\author{Francesco Periti \\
        University of Milan\\
        Via Celoria, 18 \\
        20133 Milano, Italy \\
        \texttt{francesco.periti@unimi.it} \\\And
        Nina Tahmasebi \\
        University of Gothenburg \\
        Renstr{\"o}msgatan 6\\
        40530 G{\"o}teborg, Sweden\\
        \texttt{nina.tahmasebi@gu.se}
}
  
\begin{document}
\maketitle

\begin{abstract}
Contextualized embeddings are the preferred tool for modeling Lexical Semantic Change~(LSC). Current evaluations typically focus on a specific task known as Graded Change Detection~(GCD). However, performance comparison across work are often misleading due to their reliance on diverse settings. In this paper, we evaluate state-of-the-art models and approaches for GCD under equal conditions. We further break the LSC problem into Word-in-Context (WiC) and Word Sense Induction (WSI) tasks, and compare models across these different levels. Our evaluation is performed across different languages on eight available benchmarks for LSC, and shows that (i) APD outperforms other approaches for GCD; (ii) XL-LEXEME outperforms other contextualized models for WiC, WSI, and GCD, while being comparable to GPT-4; (iii) there is a clear need for improving the modeling of word meanings, as well as focus on \textit{how}, \textit{when}, and \textit{why} these meanings change, rather than solely focusing on the extent of semantic change.
\end{abstract}

\section{Introduction}
Lexical Semantic Change (LSC) is the problem of automatically identifying words that change their meaning over time~\cite{montanelli2023survey,tahmasebi2021survey,kutuzov2018diachronic,tang2018state}. The interest in this problem has been significantly fueled by the advent of word embeddings and modern language models. After more than a decade of ad hoc evaluation, a new evaluation framework was recently introduced, aimed at assessing and comparing the performance of different models and approaches~\cite{schlechtweg2020semeval}. This framework was adopted to create benchmarks in different languages. Each benchmark includes a diachronic corpus spanning two time periods, along with a list of target words and tasks aimed at detecting word meaning change over time.  The most popular task, known as \texttt{Graded Change Detection} (GCD), consists of ranking a list of target words based on their degree of change.\vspace{8pt}

The initial excitement for word embeddings prompted researchers and practitioners to solve the GCD task by using static embedding models~\cite{schlechtweg2020semeval,shoemark2019room}. However, the shift towards more advanced Transformer architectures has established the use of contextualized embedding models as the preferred tool for addressing GCD~\cite{montanelli2023survey,kutuzov2022contextualized}. On one hand, these models distinguish the different meanings of a word by contextualizing each occurrence with a different embedding. On the other hand, the generation and processing of contextualized embeddings across entire corpora pose scalability challenges, both in terms of time and memory consumption~\cite{periti2022what,montariol2021scalable}. Different strategies have been adopted to tackle these challenges, leading to a proliferation of evaluations across diverse settings~(e.g., limited samples of benchmarks) and conditions~(e.g., pre-trained vs. fine-tuned models). As a result, these evaluations on GCD hinder a fair comparison among the performance of different models and approaches, thereby deviating from the original goal of the framework.\vspace{8pt}

Moreover, while the GCD task is attracting more and more evaluations, it addresses only a partial complexity inherent to the established framework. Notably, the framework includes three distinct aspects~\cite{schlechtweg2021dwug}: 
\begin{itemize}
    \item [\textbf{i)}] \textbf{semantic proximity judgments} of word \textit{in-context},
    \item [\textbf{ii)}] \textbf{word sense induction} based on proximity judgments,
    \item [\textbf{iii)}] \textbf{quantification of semantic change} from induced senses.
\end{itemize} 
As a matter of fact, when contextualized embedding models are used to address GCD, cosine similarities among word embeddings serve as surrogate for \textbf{(i)}, without evaluation focused on this aspect. Additionally, most approaches to GCD, pass from \textbf{(i)} to \textbf{(iii)}, sidestepping the intermediate aspect \textbf{(ii)}. That is, they quantify semantic change as overall proximity variation, without inducing word senses. Consequently, while these approaches can be evaluated through GDC, they preclude the interpretation of which meaning(s) have changed.

We argue that \textbf{(i)} and \textbf{(ii)} are equally relevant aspects as \textbf{(iii)}, constituting a fundamental aspect of the LSC problem. Their evaluation can provide valuable insights into the current state of LSC modeling, while offering a broader perspective on contextualized embedding models in Natural Language Processing (NLP).\footnote{Software is available at \url{https://github.com/FrancescoPeriti/CSSDetection}.}.

\vspace{0.3em} 
\paragraph{Original contribution of our work}
\begin{compactitem}
\vspace{0.3em}  
\item We systematically evaluate and compare various models and approaches for GCD under equal settings and conditions. Our evaluation for GCD spans eight different languages. Importantly, we perform the first evaluation over Chinese and the second evaluation for Norwegian within the existing literature. Our results show superior performance of the recent state-of-the-art model for GCD, namely XL-LEXEME, over various approaches.   
\vspace{0.3em}
\item  We are the first to evaluate contextualized embedding models for \textbf{(i)} and \textbf{(ii)} within the existing literature. Our evaluation of \textbf{(i)} and \textbf{(ii)} relies on two well-known tasks in NLP, namely \texttt{Word-in-Context} (WiC), and \texttt{Word Sense Induction} (WSI). Importantly, we evaluate various models as \textit{computational annotators}. 
\vspace{0.3em}
\item We compare GPT-4 to contextualized models through the WiC, WSI, and GCD tasks. Our evaluation reveals that GPT-4 obtains comparable performance to XL-LEXEME. In contrast to the limited accessibility\footnote{\url{https://platform.openai.com/docs/guides/rate-limits}} and high associated cost\footnote{\url{https://openai.com/pricing}} of GPT-4, XL-LEXEME is a considerably smaller, open-source model.  Thus, we argue that the use of GPT-4 is not justified for modeling the LSC problem. 
\end{compactitem}

\section{Background and related work}
\begin{figure}
   \centering
   \includegraphics[width=\columnwidth]{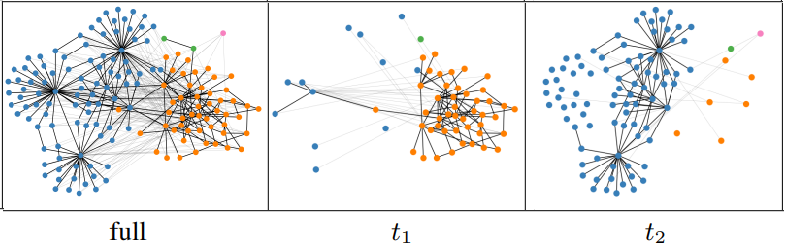}
   \caption{DWUG for the German word \textit{Eintagsfliege}. Nodes represent word usages. Edges represent the relatedness between usages. Colors indicate clusters (senses) inferred from the full graph~\cite{laicher2021explaining}.}%
 \label{fig:wug}
\end{figure}

The established LSC framework adheres to the novel annotation paradigm for word senses and encompasses \textbf{(i-iii)}~\cite{schlechtweg2021dwug}. \textbf{(i)}~Human annotators provide semantic proximity judgments for pairs of word usages \textit{sampled} from a diachronic corpus spanning two time periods. \textbf{(ii)} Word usages and judgments are represented as nodes and edges in a weighted, \textit{diachronic} graph, known as Diachronic Word Usage Graph (DWUG). This graph is then clustered with a graph clustering algorithm and the resulting clusters are interpreted as word senses (see Figure~\ref{fig:wug}), thus sidestepping the need for explicit word sense definitions. 
Finally, \textbf{(iii)} given a word, a ground truth score of semantic change is computed by comparing the probability distributions of clusters in different time periods, e.g., a cluster with most of its usages from one time period indicates a substantial semantic change. 

Originally, the framework was proposed in a shared task at SemEval-2020, including benchmarks for four languages, namely English (EN), German (DE), Swedish (SW), and Latin (LA)~\cite{schlechtweg2020semeval}. Benchmarks for Italian~\cite{basile2020diacr}, Russian (RU)~\cite{kutuzov2021rushifteval,kutuzov2021three}, Spanish (SP)~\cite{zamora2022black}, Norwegian (NO)~\cite{kutuzov2022nordiachange}, and Chinese (ZH) ~\cite{chen2023chiwuggraph,chen2022lexicon} have recently been introduced. Each benchmark\footnote{See \url{https://github.com/ChangeIsKey/LSCDBenchmark} for a comprehensive overview of available benchmarks} consists of a diachronic corpus and a set of target words over which the human annotation was conducted. The evaluation over a benchmark is typically conducted through the GCD task where the goal is to rank the targets by degree of semantic change across the corpus. The Spearman correlation between \textit{predicted} and \textit{ground truth} scores is used to evaluate models and approaches. 

\setlength{\belowcaptionskip}{-35pt}

\subsection{Approaches to Graded Change Detection}
GCD is typically addressed using two kinds of approaches for modeling word meanings:  \textit{form-} and \textit{sense}-based~\citep{montanelli2023survey,giulianelli2020analysing}. The former capture signals of change by analysing how the dominant meaning, or the degree of polysemy of a word, changes over time (e.g., \citealp{giulianelli2020analysing,martinc2020leveraging}). The latter cluster word usages according to their meanings and then estimate the semantic change of a word by comparing the cluster distribution of its usages over time (e.g., \citealp{periti2023word,martinc2020capturing}). Form- and sense-based approaches can be further distinguished into \textit{supervised}, which leverage external knowledge (e.g., dictionaries,~\citealp{rachinskiy2022black}) or other forms of supervision (e.g., Word-in-Context datasets,~\citealp{cassotti2023xl}), and \textit{unsupervised}, which rely solely on the knowledge encoded in pre-trained models (e.g.,~\citealp{aida2023textitswap}).

\subsection{Comparison of approaches}
Models and approaches for GCD have been evaluated under different settings and conditions. For example, some studies utilized the \textit{entire} diachronic corpus to estimate the change of each target (e.g.,~\citealp{periti2022what}), while others relied on smaller \textit{samples} (e.g.,~\citealp{rodina2021elmo}), or solely on the annotated word usages (e.g.,~\citealp{laicher2021explaining}). Also, different versions of the ground truth are used (e.g.,~\citealp{schlechtweg2022dwugen}). In the current literature, some studies fine-tune the models on the corpus (e.g.,~\citealp{rosin2022time}), while others directly use pre-trained models (e.g.,~\citealp{kudisov2022black}). Performance comparison are conducted across different models such as BERT (e.g.,~\citealp{laicher2021explaining}), mBERT (e.g.,~\citealp{beck2020diasense}), and XLM-R (e.g.,~\citealp{giulianelli2022fire}). However, even when the same model is employed, different layer aggregations are used, such as concatenating the output of the last four encoder layers (e.g.,~\citealp{kanjirangat2020sst}), or summing the output of all the encoder layers (e.g.,~\citealp{giulianelli2022fire}). Moreover, sense-based approaches are compared with different clustering algorithms such as Affinity Propagation (e.g.,~\citealp{martinc2020capturing}), A Posteriori affinity Propagation (e.g.,~\citealp{periti2022what}), and K-Means (e.g.,~\citealp{montariol2021scalable}). 

As a results, comparing Spearman correlation across different evaluations is often \textbf{misleading}. 
\setlength{\belowcaptionskip}{0pt}
\subsection{Current modeling of LSC}
Current modeling of LSC overlooks the procedure \textbf{(i-iii)} used to generate the ground truth. Mostly, only \textbf{(iii)} is evaluated by relying on form-based approaches. However, these approaches capture only the \textit{degree} of semantic change, preventing its interpretation. Sense-based approaches could fill this gap by explaining \textit{how} and \textit{what} has changed, but currently suffer from lower performance on \textbf{(iii)} and are therefore less pursued. As a results, it is not clear which meanings these models and approaches are capturing. There is thus a need to carefully evaluate their ability in both \textbf{(i)} and \textbf{(ii)}. 

Thus far, this evaluation is missing. To the best of our knowledge, only~\citet{laicher2021explaining} evaluate \textbf{(ii)} through the WSI task. This evaluation needs to be extended beyond a single model, using the same procedure used to generate the ground truth.

\paragraph{A systematic comparison} under equal settings and conditions is necessary to evaluate different models and approaches. Thus, we first evaluate standard form- and sense-based approaches to provide a fair performance comparison on GCD across eight languages. We then assess different models as \textit{computational annotators} by evaluating them on \textbf{(i-iii)} through WiC, WSI, and GCD. Aligning with~\citet{karjus2023machineassisted}, if computational models perform close to human-level, their usage would represent an unprecedented opportunity to scale up semantic change studies in the humanities and social sciences. 

\section{Evaluation setup}
We consider benchmarks for eight different languages: EN, LA, DE, SV, ES, RU, NO, and ZH (see Table~\ref{tab:benchmarks}). For each benchmark, we evaluate 
four different models: BERT~\cite{devlin2019bert}, mBERT, XLM-R~\cite{conneau2020unsupervised}, and XL-LEXEME~\cite{cassotti2023xl}. Aligning with the \textit{unsupervised} nature of the LSC framework, we compare pre-trained models without performing additional fine-tuning (see Table~\ref{tab:models-evaluated}). For each model and each target word in a benchmark, we collect contextualized embeddings for all its word usages in both time periods. Specifically, we generate the sets of embeddings $\Phi^1 = \{a_1, ..., a_n\}$ and $\Phi^2 = \{b_1, ..., b_m\}$ for the word usages associated to time periods $t_1$ and $t_2$, respectively. 


\subsection{Standard Graded Change Detection}\label{sec:gcd-approaches}
We compare the use of different models with four standard approaches to GCD, specifically two form-based and two sense-based. Similar to~\citet{laicher2021explaining}, we consider the raw data originally used to derive ground truth scores, instead of considering the associated corpora. This ensures an accurate evaluation under a controlled setting.

\subsection{Computational annotators}
We assess different models as computational annotators by using cosine similarities between embeddings as a surrogate of human judgments. In our evaluation, we consider word usage pairs where human judgments are available, instead of considering all potential usage pairs (as in Section~\ref{sec:gcd-approaches}). Specifically, we adhere to the framework \textbf{(i-iii)} and evaluate different models through the WiC, WSI, and GCD tasks.

Inspired by~\citet{periti2024chatgpt,laskar2023systematic,kocon2023chatgpt,karjus2023machineassisted}, we evaluate GPT-4 and compare its use to contextualized models. However, the limited accessibility and high associated cost constraint our extension only to the EN benchmark.


\section{Comparing approaches for GCD}
We evaluate different approaches for GCD using Spearman correlation between computational predictions and ground truth scores. Specifically, we process the embeddings of each target using the following approaches. 

\subsection{Form-based approaches} 
In the most recent survey on LSC by~\citet{montanelli2023survey}, it was observed that cosine distance over word prototype (PRT) and the average pairwise distance (APD) consistently demonstrated superior performance compared to alternative approaches. Thus, we employ these approaches: 

\paragraph{PRT} computes the degree of change of a word $w$ as the cosine distance between the average embeddings $\mu_1$ and $\mu_2$ (also know as \textit{prototype} embeddings) of $w$ in the time periods $t_1$ and $t_2$~\cite{martinc2020leveraging,kutuzov2020uio}. Formally, given a word $w$, we compute its degree of change by computing: 
\begin{equation}
    \text{PRT}(\Phi^1, \Phi^2) = 1 - cosine(\mu_1, \mu_2)
\end{equation} 
The intuition behind PRT is that a prototype embedding encodes 
the dominant meaning of a word, and as such, the semantic change is computed as a shift in the dominant meaning over time.

\paragraph{APD} computes the degree of change of a word $w$ as the average pairwise distance between the word embeddings in $\Phi^1$ and $\Phi^2$~\cite{giulianelli2020analysing,kutuzov2020uio}. Formally, given a word $w$, we compute its degree of change, where $d$ is cosine distance,  as follows: 
\begin{equation}
    \text{APD}(\Phi^1, \Phi^2) = \frac{1}{|\Phi^1||\Phi^2|} \ \cdot 
    \sum_{a\in \ \Phi^1, \ b \in \ \Phi^2} d(a, b)
\end{equation}
The intuition behind APD is that different word embeddings encode the polysemy of a word, and as such, the semantic change is computed as a shift in the word's degree of polysemy. 

\subsection{Sense-based approaches}
We choose two state-of-the-art sense-based approaches~\cite{montanelli2023survey}.
The first utilizes the unsupervised clustering algorithm Affinity Propagation (AP) combined with the Jensen Shannon divergence (JSD). Additionally, we employ the evolutionary extension of Affinity Propagation, called A Posteriori affinity Propagation (APP), combined with the average pairwise distances between sense prototypes (APDP). This approach is called WiDiD~\cite{periti2022what}.

\paragraph{AP+JSD} leverages the AP clustering to distinguish the different contextual usages of a given word $w$. Specifically, the embeddings $\Phi^1$, and $\Phi^2$ are \textit{collectively} clustered to generate clusters comprising embeddings from both time periods (i.e., $t_1$ and $t_2$), or embeddings exclusive from a time period (i.e., $t_1$ or $t_2$). The semantic change of $w$ is computed as the JSD between the probability distributions $p_1$ and $p_2$ of clusters in time periods $t_1$ and $t_2$. These distributions represent the relative number of embeddings from $\Phi^1$ and $\Phi^2$ grouped in each cluster, respectively~\cite{martinc2020capturing,martinc2020discovery}. Formally, 
the degree of semantic change is: 
\begin{equation}
    \text{JSD}(p_1, p_2) = \frac{1}{2}\left(KL(p_1||M) + KL(p_2||M)\right) 
\end{equation}
where $KL$ stands for Kullback-Leibler divergence and $M=\frac{(p^1+p^2)}{2}$. 
The intuition behind AP+JSD is that different clusters encode nuanced word meanings, and as such, the semantic change is computed as an overall measure of the differences in the prominence of each sense over time. 

\paragraph{WiDiD} leverages the APP clustering to distinguish the usages of a given word $w$. Specifically, the embeddings $\Phi^1$, and $\Phi^2$ are \textit{individually} clustered to generate incremental clusters of embeddings that evolve with each clustering iteration. The semantic change of $w$ is computed as the average pairwise distances between the \textit{sense prototypes} $\Psi^1$ and $\Psi^2$ of $w$ in the time periods $t_1$ and $t_2$, where $\Psi^1$ and $\Psi^2$ are the set of embeddings obtained by averaging the embeddings $\Phi^1$ and $\Phi^2$ in each cluster, respectively~\cite{periti2023word,kashleva2022black}. Formally, given a word $w$, the degree of semantic change is computed as follows:\footnote{Following~\citet{periti2023word}, we use the Canberra distance instead of the cosine distance} 
\begin{equation}
    \text{APDP}(\Phi^1, \Phi^2) = \text{APD}(\Psi^1, \Psi^2)
\end{equation}
The intuition behind WiDiD is similar to AP+JSD. However, while the latter considers change as the difference between the amount of probability for a sense over time, WiDiD is similar to APD in computing the shift in prototypical word meanings.

\begin{table*}[!ht]
\centering
\resizebox{\textwidth}{!}{%
\begin{tabular}{lcccccccccccccc}
 &  &  & \textbf{EN} & \textbf{LA} & \textbf{DE} & \textbf{SV} & \textbf{ES} & \multicolumn{3}{c}{\textbf{RU}} & \multicolumn{2}{c}{\textbf{NO}} & \textbf{ZH} & \textbf{Avg$_w$}\\ \cline{4-15} 
 &  & \multicolumn{1}{c|}{} & \multicolumn{1}{c|}{\textit{$C_1 - C_2$}} & \multicolumn{1}{c|}{\textit{$C_1 - C_2$}} & \multicolumn{1}{c|}{\textit{$C_1 - C_2$}} & \multicolumn{1}{c|}{\textit{$C_1 - C_2$}} & \multicolumn{1}{c|}{\textit{$C_1 - C_2$}} & \textit{$C_1 - C_2$} & \textit{$C_2 - C_3$} & \multicolumn{1}{c|}{\textit{$C_1 - C_3$}} & \textit{$C_1 - C_2$} & \multicolumn{1}{c|}{\textit{$C_2 - C_3$}} & \multicolumn{1}{c|}{\textit{$C_1 - C_2$}} & \multicolumn{1}{c|}{\textit{$C_i - C_j$}} \\ \cline{2-15} 
\multicolumn{1}{l|}{\multirow{2}{*}{\rotatebox{90}{\textbf{form-based} \hspace{12pt}}}} & \multicolumn{1}{c|}{APD} & \multicolumn{1}{c|}{\begin{tabular}[c]{@{}c@{}} BERT\\ mBERT\\ XLM-R\\ XL-LEXEME\\ 
\textit{SOTA: sup}.\\
\textit{SOTA: uns.}\\ \end{tabular}} & \multicolumn{1}{c|}{\begin{tabular}[c]{@{}c@{}}.563\\ .363\\ .444\\ \textbf{.886}*\\
\textit{.757}\\
\textit{.706}\\\end{tabular}} & \multicolumn{1}{c|}{\begin{tabular}[c]{@{}c@{}}-\\ .102\\ .151\\ \textbf{.231}\\\textit{-.056}\\\textit{.443}\end{tabular}} & \multicolumn{1}{c|}{\begin{tabular}[c]{@{}c@{}}.271\\ .398\\ .264\\ \textbf{.839}*\\
\textit{.877}\\
\textit{.731}\\\end{tabular}} & \multicolumn{1}{c|}{\begin{tabular}[c]{@{}c@{}}.270\\ .389\\ .257\\ \textbf{.812}*\\
\textit{.754}\\
\textit{.602}\\\end{tabular}} & \multicolumn{1}{c|}{\begin{tabular}[c]{@{}c@{}}.335 \\ .341 \\ .386 \\ \textbf{.665}*\\
\textit{n.a.}\\
\textit{n.a.}\\\end{tabular}} & \begin{tabular}[c]{@{}c@{}}.518\\ .368\\ .290\\ \textbf{.796}*\\
\textit{.799}\\
\textit{.372}\\\end{tabular} & \begin{tabular}[c]{@{}c@{}}.482\\ .345\\ .287\\ \textbf{.820}*\\
\textit{.833}\\
\textit{.480}\\\end{tabular} & \multicolumn{1}{c|}{\begin{tabular}[c]{@{}c@{}}.416\\ .386\\ .318\\ \textbf{.863}*\\
\textit{.842}\\
\textit{.457}\\\end{tabular}} & \begin{tabular}[c]{@{}c@{}}.441\\ .279\\ .195\\ \textbf{.659}\\
\textit{.757}\\\textit{.389}
\\\end{tabular} & \multicolumn{1}{c|}{\begin{tabular}[c]{@{}c@{}}.466\\ .488\\ .379\\ \textbf{.640}*\\\textit{.757}\\\textit{.387}\end{tabular}} & \multicolumn{1}{c|}{\begin{tabular}[c]{@{}c@{}}.656\\ .689\\ .500\\ \textbf{.731}*\\\textit{n.a.}\\\textit{n.a.}\end{tabular}} & \multicolumn{1}{c|}{\begin{tabular}[c]{@{}c@{}}.449\\ .371\\ .316\\ \textbf{.751}*\\\textit{}\\\textit{}\end{tabular}} \\ \cline{2-15} 
\multicolumn{1}{l|}{} & \multicolumn{1}{c|}{PRT} & \multicolumn{1}{c|}{\begin{tabular}[c]{@{}c@{}}BERT\\ mBERT\\ XLM-R\\ XL-LEXEME\\ 
\textit{SOTA: sup}.\\
\textit{SOTA: uns.}\end{tabular}} & \multicolumn{1}{c|}{\begin{tabular}[c]{@{}c@{}}.457\\ .270\\ .411\\ \textbf{.676}\\\textit{.531}\\\textit{.467}\end{tabular}} & \multicolumn{1}{c|}{\begin{tabular}[c]{@{}c@{}}-\\ .380\\ .424\\ \textbf{.506}*\\\textit{n.a.}\\\textit{.561}\end{tabular}} & \multicolumn{1}{c|}{\begin{tabular}[c]{@{}c@{}}.422\\ .436\\ .369\\ \textbf{.824}\\\textit{n.a.}\\\textit{.755}\end{tabular}} & \multicolumn{1}{c|}{\begin{tabular}[c]{@{}c@{}}.158\\ .193\\ .020\\ \textbf{.696}\\\textit{n.a.}\\\textit{.392}\end{tabular}} & \multicolumn{1}{c|}{\begin{tabular}[c]{@{}c@{}}.413\\ .543\\ .505\\
\textbf{.632}\\ \textit{n.a.}\\\textit{n.a.}\end{tabular}} & \begin{tabular}[c]{@{}c@{}}.400\\ .391\\ .321\\ \textbf{.704}\\\textit{n.a.}\\\textit{.294}\end{tabular} & \begin{tabular}[c]{@{}c@{}}.374\\ .356\\ .443\\ \textbf{.750}\\\textit{n.a.}\\\textit{313}\end{tabular} & \multicolumn{1}{c|}{\begin{tabular}[c]{@{}c@{}}.347\\ .423\\ .405\\ \textbf{.727}\\\textit{n.a.}\\\textit{313}\end{tabular}} & \begin{tabular}[c]{@{}c@{}}.507\\ .219\\ .387\\ \textbf{.764}*\\\textit{n.a.}\\\textit{.378}\end{tabular} & \multicolumn{1}{c|}{\begin{tabular}[c]{@{}c@{}}.444\\ .438\\ .149\\ \textbf{.519}\\\textit{n.a.}\\\textit{.270}\end{tabular}} & \multicolumn{1}{c|}{\begin{tabular}[c]{@{}c@{}}\textbf{.712}\\ .524\\ .558\\ .699\\\textit{n.a.}\\\textit{n.a.}\end{tabular}} & \multicolumn{1}{c|}{\begin{tabular}[c]{@{}c@{}}.406\\ .395\\ .381\\ \textbf{.693}\\\textit{}\\\textit{}\end{tabular}} \\ \cline{2-15} 
\multicolumn{1}{l|}{\multirow{2}{*}{\rotatebox{90}{\textbf{sense-based} \hspace{12pt}}}} & \multicolumn{1}{c|}{AP+JSD} & \multicolumn{1}{c|}{\begin{tabular}[c]{@{}c@{}}BERT\\ mBERT\\ XLM-R\\ XL-LEXEME\\ 
\textit{SOTA: sup}.\\
\textit{SOTA: uns.}\end{tabular}} & \multicolumn{1}{c|}{\begin{tabular}[c]{@{}c@{}}.289\\ .181\\ .278\\ \textbf{.493}\\\textit{n.a.}\\\textit{.436}\end{tabular}} & \multicolumn{1}{c|}{\begin{tabular}[c]{@{}c@{}}-\\ .277\\ \textbf{.398}\\ .033\\\textit{n.a.}\\\textit{.481}\end{tabular}} & \multicolumn{1}{c|}{\begin{tabular}[c]{@{}c@{}}.469\\ .280\\ .224\\ \textbf{.499}\\\textit{n.a.}\\\textit{.583}\end{tabular}} & \multicolumn{1}{c|}{\begin{tabular}[c]{@{}c@{}}- .090\\ .023\\  -.076\\ \textbf{.118}\\\textit{n.a.}\\\textit{.343}\end{tabular}} & \multicolumn{1}{c|}{\begin{tabular}[c]{@{}c@{}}.225\\ .067\\ .224 \\ \textbf{.392}\\\textit{n.a.}\\\textit{n.a.}\end{tabular}} & \begin{tabular}[c]{@{}c@{}}.069\\ .017\\ - .068\\ \textbf{.106}\\\textit{n.a.}\\\textit{n.a.}\end{tabular} & \begin{tabular}[c]{@{}c@{}}.279\\ .086\\ \textbf{.209}\\ .053\\\textit{n.a.}\\\textit{n.a.}\end{tabular} & \multicolumn{1}{c|}{\begin{tabular}[c]{@{}c@{}}.094\\ - .116\\ \textbf{.130}\\ .117\\\textit{n.a.}\\\textit{n.a.}\end{tabular}} & \begin{tabular}[c]{@{}c@{}}\textbf{.314}\\ .035\\ - .100\\ .297\\\textit{n.a.}\\\textit{n.a.}\end{tabular} & \multicolumn{1}{c|}{\begin{tabular}[c]{@{}c@{}}.011\\ - .090\\ .030\\ \textbf{.381}\\\textit{n.a.}\\\textit{n.a.}\end{tabular}} & \multicolumn{1}{c|}{\begin{tabular}[c]{@{}c@{}}.165\\ .465\\ .448\\ \textbf{.308}\\\textit{n.a.}\\\textit{n.a.}\end{tabular}} & \multicolumn{1}{c|}{\begin{tabular}[c]{@{}c@{}}.179\\ .077\\ .142\\ \textbf{.223}\\\textit{}\\\textit{}\end{tabular}} \\ \cline{2-15} 
\multicolumn{1}{l|}{} & \multicolumn{1}{c|}{WiDiD} & \multicolumn{1}{c|}{\begin{tabular}[c]{@{}c@{}}BERT\\ mBERT\\ XLM-R\\ XL-LEXEME\\ 
\textit{SOTA: sup}.\\
\textit{SOTA: uns.}\end{tabular}} & \multicolumn{1}{c|}{\begin{tabular}[c]{@{}c@{}}.385\\ .323\\ .564\\ \textbf{.652}\\\textit{n.a.}\\\textit{.651}\end{tabular}} & \multicolumn{1}{c|}{\begin{tabular}[c]{@{}c@{}}-\\ - .039\\ - .064\\ \textbf{.236}\\\textit{n.a.}\\\textit{-.096}\end{tabular}} & \multicolumn{1}{c|}{\begin{tabular}[c]{@{}c@{}}.355\\ .312\\ .499\\ \textbf{.677}\\\textit{n.a.}\\\textit{.527}\end{tabular}} & \multicolumn{1}{c|}{\begin{tabular}[c]{@{}c@{}}.106\\ .195\\ .129\\ \textbf{.475}\\\textit{n.a.}\\\textit{.499}\end{tabular}} & \multicolumn{1}{c|}{\begin{tabular}[c]{@{}c@{}}.383\\ .343\\ .459\\\textbf{.522}\\\textit{n.a.}\\\textit{.544}\end{tabular}} & \begin{tabular}[c]{@{}c@{}}.135\\ - .068\\ \textbf{.268}\\ .178\\\textit{n.a.}\\\textit{.273}\end{tabular} & \begin{tabular}[c]{@{}c@{}}.102\\ .160\\ .216\\ \textbf{.354}\\\textit{n.a.}\\\textit{.393}\end{tabular} & \multicolumn{1}{c|}{\begin{tabular}[c]{@{}c@{}}.243\\ .142\\ .342\\ \textbf{.364}\\\textit{n.a.}\\\textit{.407}\end{tabular}} & \begin{tabular}[c]{@{}c@{}}.233\\ .241\\ .226\\ \textbf{.561}\\\textit{n.a.}\\\textit{n.a.}\end{tabular} & \multicolumn{1}{c|}{\begin{tabular}[c]{@{}c@{}}.087\\ .290\\ .349\\ \textbf{.457}\\\textit{n.a.}\\\textit{n.a.}\end{tabular}} & \multicolumn{1}{c|}{\begin{tabular}[c]{@{}c@{}}.533\\ .338\\ .382\\ \textbf{.563}\\\textit{n.a.}\\\textit{n.a.}\end{tabular}} & \multicolumn{1}{c|}{\begin{tabular}[c]{@{}c@{}}.239\\ .181\\ .314\\ \textbf{.422}\\\textit{}\\\textit{}\end{tabular}} \\ \cline{2-15} 
\end{tabular}%
}
\caption{\textbf{Evaluation of standard approaches to GCD} in terms of Spearman correlation. Top score for each approach and benchmark in \textbf{bold}. The top score of each benchmark is marked with an asterisk (*). We include state-of-the-art performance achieved by \textit{supervised} (sup.) and \textit{unsupervised} (uns.) approaches in \textit{italic}. Avg is the weighted average score based on the number of targets in each benchmark. Results not available denoted as n.a.}
\label{tab:form-vs-sense}
\end{table*}

\subsection{Evaluation results - Table~\ref{tab:form-vs-sense}}
We present the results of our evaluation in Table~\ref{tab:form-vs-sense} for both form- and sense-based approaches. For the sake of comparison, we include state-of-the-art (SOTA) results in Table~\ref{tab:sota}.\footnote{Our comparison includes results from different benchmarks using the same approaches. However, some benchmarks might have been assessed using other approaches.}
As a general remark, we note instances where our results surpass SOTA (e.g., XL-LEXEME+APD for EN). We attribute this to the controlled setting established in our experiments. We note also instances where our results are lower than SOTA (e.g., BERT+APD for SV). This discrepancy may be influenced by various factors such as \textit{different versions} of the benchmarks (e.g., 37 vs 46 targets for EN in DWUG version 2.0.1,~\citealp{schlechtweg2020semeval}). Additionally, \textit{variations in text pre-processing} can play a beneficial role. For instance,~\citet{laicher2021explaining} demonstrate the effectiveness of lemmatization to mitigate word form biases, while~\citet{martinc2020discovery} suggest that filtering Named Entities can help models avoid inflating semantic change.  Moreover, some studies \textit{fine-tune or utilize different embedding layers}, whereas we adhere to the standard, generally adopted procedures without fine-tuning, considering embeddings generated from the last (i.e., 12$^{\text{th}}$) layer of the models. Finally, there are sometimes significantly different results reported by different studies under similar conditions. For instance, \citet{zhou2023finer} achieve a correlation of .706 using pre-trained BERT and APD, whereas others typically report correlations ranging between .400 and .600 (e.g., .489,~\citealp{keidar2022slangvolution}; .514,~\citealp{giulianelli2020analysing}; .546,~\citealp{kutuzov2020uio}; .571,~\citealp{laicher2021explaining}). This disparity cannot currently be explained.

\paragraph{Languages.} We obtain strong correlations with all benchmarks but LA. Our results show a \textit{weighted average} correlation of \textbf{.751} when employing XL-LEXEME~+~APD. 
In this calculation, we assign weights based on the number of targets in each benchmark, considering larger sets more reliable than smaller ones. For LA, it can be argued that the models were not directly tailored or fine-tuned for Latin. However, XL-LEXEME demonstrates optimal performance in GCD in SV and medium performance in SP and NO without specific training on either~\cite{cassotti2023xl}. This leads us to consider that the quality of the LA benchmark potentially is lower than other benchmarks, as it was developed using a different procedure~\cite{schlechtweg2020semeval}.

\paragraph{Form-based vs Sense-based.} We note that form-based approaches significantly outperform sense-based approaches. Our results consistently highlight APD as the most effective approach, regardless of the skewness in the distribution of judgments, as previously argued by~\citet{kutuzov2020uio}. In addition, WiDiD consistently demonstrate superior performance over AP+JSD. This can be attributed to the use of i) an evolutionary clustering algorithm, which enables to consider the time dimension of text in a dynamic way; or, alternatively ii) APD over sense-prototypes, as APD has demonstrated high effectiveness.

Our \textbf{leaderboard} is as follows: APD, PRT, WiDiD, AP+JSD. Although form-based approaches exhibit superior effectiveness, they fall short in capturing word meanings and interpreting detected semantic changes. In contrast, although sense-based approaches theoretically facilitate such modeling and interpretation, they obtain poor results in GCD, raising concerns about their reliability and whether they capture meaningful patterns or 
produce noisy aggregation. We will investigate this in Section \ref{sec:five}.

\begin{table*}[!ht]
\centering
\resizebox{\textwidth}{!}{%
\begin{tabular}{ccccccccccccc}
\multicolumn{1}{l}{} &  & \textbf{EN} & \textbf{DE} & \textbf{SV} & \textbf{ES} & \multicolumn{3}{c}{\textbf{RU}} & \multicolumn{2}{c}{\textbf{NO}} & \textbf{ZH} & \textbf{Avg$_w$} \\ \cline{3-13} 
\multicolumn{1}{l}{} & \multicolumn{1}{c|}{} & \multicolumn{1}{c|}{\textit{$C_1 - C_2$}}  & \multicolumn{1}{c|}{\textit{$C_1 - C_2$}} & \multicolumn{1}{c|}{\textit{$C_1 - C_2$}} & \multicolumn{1}{c|}{\textit{$C_1 - C_2$}} & \textit{$C_1 - C_2$} & \textit{$C_2 - C_3$} & \multicolumn{1}{c|}{\textit{$C_1 - C_3$}} & \textit{$C_1 - C_2$} & \multicolumn{1}{c|}{\textit{$C_2 - C_3$}} & \multicolumn{1}{c|}{\textit{$C_1 - C_2$}} & \multicolumn{1}{c|}{\textit{$C_i - C_j$}}\\ \cline{2-13} 
\multicolumn{1}{c|}{\rotatebox{90}{\textbf{WiC}}} & \multicolumn{1}{c|}{\begin{tabular}[c]{@{}c@{}}BERT\\ mBERT\\ XLM-R\\ XL-LEXEME\\ GPT-4.0 \\ \textit{Agreement}\end{tabular}} & \multicolumn{1}{c|}{\begin{tabular}[c]{@{}c@{}}.503\\ .332\\ .352\\ \textbf{.626}\\ .606\\ \textit{.633}\end{tabular}}  & \multicolumn{1}{c|}{\begin{tabular}[c]{@{}c@{}}.350\\ .344\\ .289\\ \textbf{.628}\\ -\\ \textit{.666}\end{tabular}} & \multicolumn{1}{c|}{\begin{tabular}[c]{@{}c@{}}.221\\ .284\\ .255\\ \textbf{.631}\\ - \\ \textit{.672}\end{tabular}} & \multicolumn{1}{c|}{\begin{tabular}[c]{@{}c@{}}.319\\ .289\\ .288\\ \textbf{.547}\\ -\\\textit{.531}\end{tabular}} & \begin{tabular}[c]{@{}c@{}}.314\\ .280\\ .212\\ \textbf{.549}\\ -\\\textit{.531}\end{tabular} & \begin{tabular}[c]{@{}c@{}}.344\\ .273\\ .250\\ \textbf{.558}\\ - \\ \textit{.567}\end{tabular} & \multicolumn{1}{c|}{\begin{tabular}[c]{@{}c@{}}.350\\ .293\\ .251\\ \textbf{.564}\\ -\\ \textit{.564}\end{tabular}} & \begin{tabular}[c]{@{}c@{}}.429\\ .283\\ .317\\ \textbf{.484}\\ - \\\textit{.761}\end{tabular} & \multicolumn{1}{c|}{\begin{tabular}[c]{@{}c@{}}.406\\ .333\\ .261\\ \textbf{.521}\\ -\\\textit{.667}\end{tabular}} & \multicolumn{1}{c|}{\begin{tabular}[c]{@{}c@{}}.516\\ .413\\ .392\\ \textbf{.630}\\ -\\\textit{.602}\end{tabular}} & \multicolumn{1}{c|}{\begin{tabular}[c]{@{}c@{}}.358\\ .301\\ .272\\ \textbf{.568}\\ -\\ \textit{.593}\end{tabular}} \\ \cline{2-13} 
\multicolumn{1}{c|}{\rotatebox{90}{\textbf{WSI}}} & \multicolumn{1}{c|}{\begin{tabular}[c]{@{}c@{}}BERT\\ mBERT\\ XLM-R\\ XL-LEXEME\\ GPT-4.0\end{tabular}} & \multicolumn{1}{c|}{\begin{tabular}[c]{@{}c@{}}.136 / .700\\ .067  / .644\\ .068 / .737\\ .273 / .834\\ \textbf{.340} / \textbf{.877}\end{tabular}}  & \multicolumn{1}{c|}{\begin{tabular}[c]{@{}c@{}}.047 / .662\\ .054 / .679\\ .024 / .725\\ \textbf{.300} / \textbf{.788}\\ - / -\end{tabular}} & \multicolumn{1}{c|}{\begin{tabular}[c]{@{}c@{}}.023 / .596\\ .024 / .648\\ .031 / .680\\ \textbf{.249} / \textbf{.766}\\ - / -\end{tabular}} & \multicolumn{1}{c|}{\begin{tabular}[c]{@{}c@{}}.189 / .695\\ .228 / .700\\ .164 / .755\\ \textbf{.400} / \textbf{.820}\\ - / -\end{tabular}} & \begin{tabular}[c]{@{}c@{}}- / -\\ - / -\\ - / -\\ - / -\\ - / -\end{tabular} & \begin{tabular}[c]{@{}c@{}}- / -\\ - / -\\ - / -\\ - / -\\ - / -\end{tabular} & \multicolumn{1}{c|}{\begin{tabular}[c]{@{}c@{}}- / -\\ - / -\\ - / -\\ - / -\\ - / -\end{tabular}} & \begin{tabular}[c]{@{}c@{}}.251 / .771\\ .241 / .759\\ .179 / .775\\ \textbf{.337} / \textbf{.806}\\ - / -\end{tabular} & \multicolumn{1}{c|}{\begin{tabular}[c]{@{}c@{}}.247 / .758\\ .159 / .753\\ .183 / .715\\ \textbf{.304} / \textbf{.808}\\ - / -\end{tabular}} & \multicolumn{1}{c|}{\begin{tabular}[c]{@{}c@{}}.279 / .759\\ .172 / .713\\ .279 / .806\\ \textbf{.448} / \textbf{.836}\\ - / -\end{tabular}} & \multicolumn{1}{c|}{\begin{tabular}[c]{@{}c@{}}.166 / .702\\ .146 / .696\\ .133 / .743\\ \textbf{.339} / \textbf{.810}\\ - / -\end{tabular}}\\ \cline{2-13}
\multicolumn{1}{c|}{\rotatebox{90}{\textbf{GCD}}} & \multicolumn{1}{c|}{\begin{tabular}[c]{@{}c@{}}BERT\\ mBERT\\ XLM-R\\ XL-LEXEME\\ GPT-4.0
\end{tabular}} & \multicolumn{1}{c|}{\begin{tabular}[c]{@{}c@{}}.425\\ .120\\ .219\\ .801\\ \textbf{.818}
\end{tabular}}  & \multicolumn{1}{c|}{\begin{tabular}[c]{@{}c@{}}.116\\ .205\\ .069\\ \textbf{.799}\\ -
\end{tabular}} & \multicolumn{1}{c|}{\begin{tabular}[c]{@{}c@{}}.148\\ .234\\ .143\\ \textbf{.721}\\ -
\end{tabular}} & \multicolumn{1}{c|}{\begin{tabular}[c]{@{}c@{}}.284\\ .394\\ .464\\ \textbf{.655}\\ -
\end{tabular}} & \begin{tabular}[c]{@{}c@{}}.487\\ .372\\ .284\\ \textbf{.780}\\ -
\end{tabular} & \begin{tabular}[c]{@{}c@{}}.452\\ .325\\ .301\\ \textbf{.824}\\ -
\end{tabular} & \multicolumn{1}{c|}{\begin{tabular}[c]{@{}c@{}}.469\\ .408\\ .375\\ \textbf{.851}\\ -
\end{tabular}} & \begin{tabular}[c]{@{}c@{}}.571\\ .290\\ .395\\ \textbf{.620}\\ -
\end{tabular} & \multicolumn{1}{c|}{\begin{tabular}[c]{@{}c@{}}.521\\ .454\\ .345\\ \textbf{.567}\\ -
\end{tabular}} & \multicolumn{1}{c|}{\begin{tabular}[c]{@{}c@{}}\textbf{.808}\\ .737\\ .557\\ .716\\ -
\end{tabular}} & \multicolumn{1}{c|}{\begin{tabular}[c]{@{}c@{}}.422\\ .357\\ .324\\ \textbf{.754}\\ -
\end{tabular}}\\ \cline{2-13}
\end{tabular}%
}
\caption{\textbf{Evaluation of contextualized models as computational annotators}: Spearman correlation  for WiC and GCD, Adjusted Random Index and Purity~(ARI~/~PUR) for WSI. 
Top score for each approach and benchmark is highlighted in \textbf{bold}. Avg is a weighted average based on the number of targets in each benchmark test set. For the sake of comparison, we report the Krippendorff's $\alpha$ score for inter-human annotator \textit{agreement} in WiC (\textit{italic}).}
\label{tab:wic-wsi-gcd}
\end{table*}

\paragraph{Supervised vs Unsupervised.} We note that the use of supervision significantly improves the modeling of semantic change for both form- and sense-based approaches. While~\citet{cassotti2023xl} have previously evaluated XL-LEXEME~+~APD, we extend the evaluation to sense-based approaches, demonstrating that \textit{supervision} enhances the performance of AP+JSD and WiDiD.

\paragraph{Models.} We note that the use of XL-LEXEME significantly improves the modeling of LSC compared to standard BERT, mBERT, and XLM-R. However, we observe a pattern in performance, indicating that on average, BERT performs better than mBERT, which, in turn, performs better than XLM-R for form-based approaches. This suggests that the use of XLM-R models is not more effective than BERT models for LSC, confirming the medium-low correlation coefficients obtained by~\citet{giulianelli2022fire} using XLM-R.

\paragraph{Layers.} As different works employ different embedding layers, we repeat our evaluation by considering embeddings generated by each layer of BERT, mBERT, and XLM-R (see Appendix~\ref{app:layers}). Our evaluation aligns with recent findings on other downstream tasks~\cite{ma2019universal,coenen2019visualizing,liang2023named} and shows that using early layers consistently results in higher performance. For example, we note a correlation  of .747 for ZH by using layer 4, compared to .656 obtained by using the last layer of BERT. On average, and in line with~\citet{periti2023time}, we find that the best results for each language are obtained by leveraging embeddings from layers 8 -- 10.

Furthermore, since previous studies aggregated outputs from different layers, we also use aggregated embeddings extracted from different layers through sum and concatenation (see Appendix~\ref{app:layers}). Specifically, our evaluation covers all possible layer combinations with lengths of 2 (e.g., layers 1 and 2), 3 (e.g., layers 6, 7, and 8), and 4 (e.g., layers 9, 10, 11, 12). We find no improvement in aggregating the output of the last four layers for addressing GCD. By employing alternative layer combinations, we obtain higher correlation compared to both the last layer and the last four layers. For instance, for EN, using the sum of layers 2, 4, 5, and 8 for APD+BERT, or the concatenation of layers 4, 5, 6, and 11 for WiDiD+BERT, results in correlation of .692 and .760, respectively; compared to .563 (APD) and .385 (WiDiD) by using the last BERT layer. However, no combination consistently emerges as the optimal choice across various benchmarks or models. Instead, we observe that using a middle layer, such as layer 8, tends to be advantageous across benchmarks and models compared to the last layer or the aggregation of the last four layers (see Figure~\ref{fig:layers} and~\ref{fig:layers1}).

\section{Computational annotation} \label{sec:five}
We evaluate different models on reproducing human judgments \textbf{(i)}, the inferred word senses \textbf{(ii)}, and the resulting change scores \textbf{(iii)}. 

We leverage models as annotators, hence the term \textit{computational annotator}, using the same procedure employed for benchmark construction~\cite{schlechtweg2023human,schlechtweg2021dwug,schlechtweg2020semeval,schlechtweg2020simulating,schlechtweg2018diachronic}. However, we cannot evaluate LA as the benchmark was developed differently nor \textbf{(ii)} for the RU benchmark since no word senses were provided~\cite{kutuzov2021rushifteval,kutuzov2021three}. 

\subsection{(i) - Word-in-Context}\label{subsec:WiC}
Given a benchmark, a word usage pair is associated with two contexts, $c_1$ and $c_2$, along with the average judgment of multiple annotators (see Example~\ref{ex:wic}). We thus use the cosine similarity between the embeddings of $w$ in the contexts $c_1$ and $c_2$ as computational proximity judgement.

Our evaluation is grounded in the Word-in-Context (WiC) task~\cite{loureiro2022tempowic,raganato2020xl,pilehvar2019wic}. In contrast to the original WiC definition, our WiC evaluation aligns with the continuous framework introduced by~\citet{armendariz2020semeval} in the Graded Word Similarity in Context task. Specifically, we evaluate the quality of computational predictions by computing the Spearman correlation with human judgments. 

\subsection{(ii) - Word Sense Induction}
We first create a DWUG using the computational annotations in Section~\ref{subsec:WiC}. Then, we derive sense clusters through a variation of correlation clustering~\cite{bansal2004correlation} on the DWUG.  

Our evaluation is grounded in the Word Sense Induction (WSI) task~\cite{aksenova2022rudsi,aksenova2022rudsi,manandhar2010semeval,agirre2007semeval}. We evaluate the quality of clusters from computationally annotated DWUGs against clusters from human-annotated DWUGs. Specifically, we use Adjusted Rand Index (ARI,~\citealp{hubert1985comparing}) and Purity (PUR,~\citealp{manning2009introduction}) as metrics to quantify the cluster agreement. ARI comprehensively evaluates the similarity among clustering result. However, it may yield low scores when a clustering result contains numerous small, yet coherent clusters. This does not necessarily indicate poor clustering quality, especially when the clusters are semantically meaningful. PUR assigns each cluster to the class that is most frequent in the cluster, measuring the accuracy of this assignment by counting the relative number of correctly assigned elements.

\subsection{(iii) - Graded Change Detection}
Given a word $w$, we split its DWUG into two subgraphs representing nodes from the two time periods (see Figure~\ref{fig:wug}) and quantify the semantic change of $w$ by computing the $\sqrt{JSD}$ between the two time-specific cluster distributions. In contrast, for RU, we adhere to the RuShiftEval procedure and quantify semantic change through the application of the COMPARE metric that directly measures the mean relatedness of annotated word usage pairs as semantic change scores~\cite{schlechtweg2018diachronic}. Our evaluation is based on the GCD task and thus use Spearman correlation as evaluation metric between predicted ranking and ground truth rankings.

\subsection{Evaluation results -- Table~\ref{tab:wic-wsi-gcd}}

\paragraph{(i) - Word-in-Context} Our evaluation reveals that pre-trained models such as BERT, mBERT, and XLM-R demonstrate a 
low average correlation with human judgments (.358, .301, .272). In contrast, XL-LEXEME and GPT-4 emerge as powerful solutions for scaling up and aiding human annotations. For EN, they obtain a moderately strong correlation (.626, .606) with human judgments, only marginally lower than the Krippendorf $\alpha$ human agreement (.633). In particular, XL-LEXEME slightly outperforms a considerably larger model like GPT-4 in terms of parameters, at a considerable lower cost. In contrast to previous cross-lingual evaluation~\cite{conneau2020unsupervised} and in line with the finding in Table~\ref{tab:form-vs-sense}, mBERT consistently outperforms XLM-R. However, our results highlight the advantageous use of monolingual BERT models over the multilingual ones, for assessing \textbf{(i)} - WiC. 

We consider the WiC evaluation to be the most valuable as it involves a direct comparison between computational predictions and human judgments.

\paragraph{(ii) - Word Sense Induction} Our evaluation indicates that moderate performance in \textbf{(i)}-WiC leads to moderately \textit{low} performance in inferring word sense. We obtain low ARI scores across all models and benchmarks, with XL-LEXEME and GPT-4 exhibiting the highest values. Specifically, GPT-4 outperforms XL-LEXEME (with .340 compared to .273) in ARI for EN. However, we highlight that even such low scores represent a moderately \textit{high} result, given an inter-annotator agreement of .633.

XL-LEXEME consistently demonstrates high PUR scores across all benchmarks, while other models yield slightly lower PUR scores, suggesting that some word sense patterns are captured when using contextualized models. Previous studies highlight that contextualized models tend to produce a large number of clusters~\cite{martinc2020capturing,periti2022what}, thereby influencing PUR scores. Therefore, it is crucial to interpret PUR in conjunction with ARI.

\paragraph{(iii) - Graded Change Detection}
As for GCD, we obtain average results for BERT, mBERT, XLM-R, and XL-LEXEME equal to .422, .357, .324, .754, respectively. These results are consistent with those presented in Table~\ref{tab:form-vs-sense}, when compared to form-based approaches (.316 -- .751). We observe that employing more word usage pairs, as in Table~\ref{tab:form-vs-sense}, proves beneficial for certain benchmarks in the GCD tasks (e.g., XL-LEXEME+APD for EN and DE). However, we note that these results for \textbf{(ii)} - WSI are significantly higher to those obtained by sense-based approaches (.077 -- .422). This can likely be attributed the fact that here we are using the same clustering algorithm that was used for obtaining the ground truth clusters, or to the fact that the clustering algorithm is more able to capture nuanced word meaning than AP and APP. In contrast, for RU, following the RuShiftEval procedure does not improve the performance and results between Table~\ref{tab:form-vs-sense} and~\ref{tab:wic-wsi-gcd} are somewhat comparable.

\section{Concluding remarks}
We have performed a first-ever evaluation of models and approaches for modeling 
LSC under equal settings and conditions, over eight different languages. First, we evaluated different models combined with standard approaches to the popular GCD task. 
In particular, we consider BERT, mBERT, XLM-R, XL-LEXEME as pre-trained models, APD and PRT as form-based approaches, and AP+JSD and WiDiD as sense-based approaches. We find that the XL-LEXEME  consistently outperforms other models across all approaches, and thus should be used as the defacto standard. We also find that form-based approaches significantly outperform sense-based approaches, with APD as the best approach for GCD. Among the sense-based approaches, we find that \textit{evolutionary} clustering is advantageous in contrast to static clustering and should be a focus of future work. 
We additionally extended the evaluation to includes the WiC and WSI tasks, 
both inherently crucial to solve the complex task of LSC. 
We compare GPT-4 to the previous models and find that GPT-4 and XL-LEXEME both perform close to human-level 
while the other models obtain only low-moderate performance. Due to the costs associated with using GPT-4, it is not affordable to evaluate it on the remaining languages. Since XL-LEXEME obtains results close those of GPT-4, even beating it for the WiC task, we argue that XL-LEXEME can be used for LSC tasks as a affordable, scalable solution.

All in all, considering the current state of the LSC modeling, we argue that 
\textbf{only obtaining state-of-the-art performance on GCD does not solve the LSC problem}, as there is a clear need to \textbf{distinguish  the different senses of a word and how these evolve over time}~\cite{periti2023word}.  GCD maintains 
relevance for identifying words that have changed across multiple time periods 
in need of further \textit{sense-based} modeling. GCD also serves to quantify the 
change on the level of vocabulary. 
In conclusion, we offer a first comparable evaluation of contextualized word embeddings for LSC and establish clear settings that should be used for future comparison and evaluation. With this work, we want to raise awareness of the current trend of the community in modeling only the GCD task. Our aim is to shift the focus from merely assessing \textit{how much} to \textit{how}, \textit{when}, and \textit{why}, prompting the development of both \textit{unsupervised} and \textit{supervised} approaches for addressing the full spectrum of LSC. 


\section{Limitations}
There are limitations we had to consider in the making of this paper. Firstly, we could not evaluate GPT-4 across all languages due to both price and API limitations. This means that while the results are comparable with XL-LEXEME for EN, we do not know how GPT-4 will behave for the other languages. Although we are aware of open source solution such as LLaMA, our initial experiments,  revealed that its performance does not match that of GPT-4. As LLaMA still necessitates expensive research infrastructure, we chose to focus only GPT-4. Our decision to use GPT-4 over the cheaper GPT-3 is based on recent studies showing conflicting results across different tasks. Notably, \citet{karjus2023machineassisted} reported high scores for GPT-4 in the GCD task. However, \citet{periti2024chatgpt,laskar2023systematic,kocon2023chatgpt} reported low scores for the WiC task when employing GPT-3. As a result, we opted for GPT-4 to ensure relevance and accuracy in our evaluations.

In this paper, we evaluate different contextualized models utilizing the popular Transformers library for deep learning maintained by Hugging Face~\cite{wolf2020transformers}. We specifically excluded the evaluation of a BERT model for Latin, opting instead to focus on mBERT, XLM-R, and XL-LEXEME. At the beginning of our evaluation, we were not aware of any experiments using Latin BERT models to address GCD, nor were we aware of an open BERT version for Latin on the Hugging Face platform. As we have only recently become aware of novel BERT models that are exclusively trained and fine-tuned for Latin~\cite{riemenschneider2023exploring,lendvai2022finetuning}, we plan to further test and utilize these models in our future work.

To make a fair comparison between different contextualized models, we employed the same procedure across all benchmarks and languages. However, different languages have different structures and hence different requirements. It would be equally fair to have different processing of the different benchmarks (e.g., lemmatization for German, \citealp{laicher2021explaining}). We opted to reduce the number of open variables to be able to make this first evaluation. Future work could optimize each language and then compare model performance. 

Lastly, the models compared in this study, despite sharing similar architectures, tokenize text sequences differently based on their reference vocabulary. Consequently, a word may be split into different subtokens by one model and represented as a single token by another. Additionally, when contexts exceed the maximum input size, different models may truncate them at various points. Adhering to standard procedures in the field of LSC, we use the average embeddings of sub-words when a word is split into multiple sub-words. However, the impact of different truncation methods was not evaluated.

\section*{Acknowledgments}
This work has in part been funded by the project Towards Computational Lexical Semantic Change Detection supported by the Swedish Research Council (2019–2022; contract 2018-01184), and in part by the research program Change is Key! supported by Riksbankens Jubileumsfond (under reference number M21-0021). The computational resources were provided by the National Academic Infrastructure for Supercomputing in Sweden (NAISS).

\bibliography{bibtex}

\begin{thebibliography}{75}
\expandafter\ifx\csname natexlab\endcsname\relax\def\natexlab#1{#1}\fi

\bibitem[{Agirre and Soroa(2007)}]{agirre2007semeval}
Eneko Agirre and Aitor Soroa. 2007.
\newblock \href {https://aclanthology.org/S07-1002} {{SemEval-2007 Task 02: Evaluating Word Sense Induction and Discrimination Systems}}.
\newblock In \emph{Proceedings of the Fourth International Workshop on Semantic Evaluations ({S}em{E}val-2007)}, pages 7--12, Prague, Czech Republic. Association for Computational Linguistics.

\bibitem[{Aida and Bollegala(2023)}]{aida2023textitswap}
Taichi Aida and Danushka Bollegala. 2023.
\newblock \href {https://doi.org/10.18653/v1/2023.findings-emnlp.520} {{Swap and Predict -- Predicting the Semantic Changes in Words across Corpora by Context Swapping}}.
\newblock In \emph{Findings of the Association for Computational Linguistics: EMNLP 2023}, pages 7753--7772, Singapore. Association for Computational Linguistics.

\bibitem[{Aksenova et~al.(2022)Aksenova, Gavrishina, Rykov, and Kutuzov}]{aksenova2022rudsi}
Anna Aksenova, Ekaterina Gavrishina, Elisei Rykov, and Andrey Kutuzov. 2022.
\newblock \href {https://aclanthology.org/2022.textgraphs-1.9} {{RuDSI: Graph-based Word Sense Induction Dataset for Russian}}.
\newblock In \emph{Proceedings of TextGraphs-16: Graph-based Methods for Natural Language Processing}, pages 77--88, Gyeongju, Republic of Korea. Association for Computational Linguistics.

\bibitem[{Armendariz et~al.(2020)Armendariz, Purver, Pollak, Ljube{\v{s}}i{\'c}, Ul{\v{c}}ar, Vuli{\'c}, and Pilehvar}]{armendariz2020semeval}
Carlos~Santos Armendariz, Matthew Purver, Senja Pollak, Nikola Ljube{\v{s}}i{\'c}, Matej Ul{\v{c}}ar, Ivan Vuli{\'c}, and Mohammad~Taher Pilehvar. 2020.
\newblock \href {https://doi.org/10.18653/v1/2020.semeval-1.3} {{SemEval-2020 Task 3: Graded Word Similarity in Context}}.
\newblock In \emph{Proceedings of the Fourteenth Workshop on Semantic Evaluation}, pages 36--49, Barcelona (online). International Committee for Computational Linguistics.

\bibitem[{Bansal et~al.(2004)Bansal, Blum, and Chawla}]{bansal2004correlation}
Nikhil Bansal, Avrim Blum, and Shuchi Chawla. 2004.
\newblock \href {https://doi.org/doi.org/10.1023/B:MACH.0000033116.57574.95} {{Correlation clustering}}.
\newblock \emph{Machine learning}, 56:89--113.

\bibitem[{Basile et~al.(2020)Basile, Caputo, Caselli, Cassotti, and Varvara}]{basile2020diacr}
Pierpaolo Basile, Annalina Caputo, Tommaso Caselli, Pierluigi Cassotti, and Rossella Varvara. 2020.
\newblock \href {https://ceur-ws.org/Vol-2765/paper158.pdf} {{DIACR-Ita@ EVALITA2020: Overview of the EVALITA2020 DiachronicLexical Semantics (DIACR-Ita) Task}}.
\newblock In \emph{Proceedings of the Evaluation Campaign of Natural Language Processing and Speech Tools for Italian (EVALITA)}, Online. CEUR-WS.

\bibitem[{Beck(2020)}]{beck2020diasense}
Christin Beck. 2020.
\newblock \href {https://doi.org/10.18653/v1/2020.semeval-1.4} {{DiaSense at SemEval-2020 Task 1: Modeling Sense Change via Pre-trained BERT Embeddings}}.
\newblock In \emph{Proceedings of the Fourteenth Workshop on Semantic Evaluation}, pages 50--58, Barcelona (online). International Committee for Computational Linguistics.

\bibitem[{Cassotti et~al.(2023)Cassotti, Siciliani, DeGemmis, Semeraro, and Basile}]{cassotti2023xl}
Pierluigi Cassotti, Lucia Siciliani, Marco DeGemmis, Giovanni Semeraro, and Pierpaolo Basile. 2023.
\newblock \href {https://doi.org/10.18653/v1/2023.acl-short.135} {{XL-LEXEME: WiC Pretrained Model for Cross-Lingual LEXical sEMantic changE}}.
\newblock In \emph{Proceedings of the 61st Annual Meeting of the Association for Computational Linguistics (Volume 2: Short Papers)}, pages 1577--1585, Toronto, Canada. Association for Computational Linguistics.

\bibitem[{Chen et~al.(2022)Chen, Chersoni, and Huang}]{chen2022lexicon}
Jing Chen, Emmanuele Chersoni, and Chu-ren Huang. 2022.
\newblock \href {https://doi.org/10.18653/v1/2022.lchange-1.11} {{Lexicon of Changes: Towards the Evaluation of Diachronic Semantic Shift in Chinese}}.
\newblock In \emph{Proceedings of the 3rd Workshop on Computational Approaches to Historical Language Change}, pages 113--118, Dublin, Ireland. Association for Computational Linguistics.

\bibitem[{Chen et~al.(2023{\natexlab{a}})Chen, Chersoni, Schlechtweg, Prokic, and Huang}]{chen2023chiwuggraph}
Jing Chen, Emmanuele Chersoni, Dominik Schlechtweg, Jelena Prokic, and Chu-Ren Huang. 2023{\natexlab{a}}.
\newblock \href {https://aclanthology.org/2023.lchange-1.10} {{ChiWUG: A Graph-based Evaluation Dataset for {C}hinese Lexical Semantic Change Detection}}.
\newblock In \emph{Proceedings of the 4th Workshop on Computational Approaches to Historical Language Change}, pages 93--99, Singapore. Association for Computational Linguistics.

\bibitem[{Chen et~al.(2023{\natexlab{b}})Chen, Chersoni, Schlechtweg, Prokic, and Huang}]{chen2023chiwug}
Jing Chen, Emmanuele Chersoni, Dominik Schlechtweg, Jelena Prokic, and Chu-Ren Huang. 2023{\natexlab{b}}.
\newblock \href {https://doi.org/10.18653/v1/2023.lchange-1.10} {{ChiWUG: A Graph-based Evaluation Dataset for Chinese Lexical Semantic Change Detection}}.
\newblock In \emph{Proceedings of the 4th Workshop on Computational Approaches to Historical Language Change}, pages 93--99, Singapore. Association for Computational Linguistics.

\bibitem[{Conneau et~al.(2020)Conneau, Khandelwal, Goyal, Chaudhary, Wenzek, Guzm{\'a}n, Grave, Ott, Zettlemoyer, and Stoyanov}]{conneau2020unsupervised}
Alexis Conneau, Kartikay Khandelwal, Naman Goyal, Vishrav Chaudhary, Guillaume Wenzek, Francisco Guzm{\'a}n, Edouard Grave, Myle Ott, Luke Zettlemoyer, and Veselin Stoyanov. 2020.
\newblock \href {https://doi.org/10.18653/v1/2020.acl-main.747} {{Unsupervised Cross-lingual Representation Learning at Scale}}.
\newblock In \emph{Proceedings of the 58th Annual Meeting of the Association for Computational Linguistics}, pages 8440--8451, Online. Association for Computational Linguistics.

\bibitem[{Devlin et~al.(2019)Devlin, Chang, Lee, and Toutanova}]{devlin2019bert}
Jacob Devlin, Ming-Wei Chang, Kenton Lee, and Kristina Toutanova. 2019.
\newblock \href {https://doi.org/10.18653/v1/N19-1423} {{BERT: Pre-training of Deep Bidirectional Transformers for Language Understanding}}.
\newblock In \emph{Proceedings of the 2019 Conference of the North {A}merican Chapter of the Association for Computational Linguistics: Human Language Technologies, Volume 1 (Long and Short Papers)}, pages 4171--4186, Minneapolis, Minnesota. Association for Computational Linguistics.

\bibitem[{Giulianelli et~al.(2020)Giulianelli, Del~Tredici, and Fern{\'a}ndez}]{giulianelli2020analysing}
Mario Giulianelli, Marco Del~Tredici, and Raquel Fern{\'a}ndez. 2020.
\newblock \href {https://doi.org/10.18653/v1/2020.acl-main.365} {{Analysing Lexical Semantic Change with Contextualised Word Representations}}.
\newblock In \emph{Proceedings of the 58th Annual Meeting of the Association for Computational Linguistics}, pages 3960--3973, Online. Association for Computational Linguistics.

\bibitem[{Giulianelli et~al.(2022)Giulianelli, Kutuzov, and Pivovarova}]{giulianelli2022fire}
Mario Giulianelli, Andrey Kutuzov, and Lidia Pivovarova. 2022.
\newblock \href {https://doi.org/10.18653/v1/2022.lchange-1.6} {{Do Not Fire the Linguist: Grammatical Profiles Help Language Models Detect Semantic Change}}.
\newblock In \emph{Proceedings of the 3rd Workshop on Computational Approaches to Historical Language Change}, pages 54--67, Dublin, Ireland. Association for Computational Linguistics.

\bibitem[{Hubert and Arabie(1985)}]{hubert1985comparing}
Lawrence Hubert and Phipps Arabie. 1985.
\newblock \href {https://doi.org/doi.org/10.1007/BF01908075} {{Comparing Partitions}}.
\newblock \emph{Journal of classification}, 2:193--218.

\bibitem[{Kanjirangat et~al.(2020)Kanjirangat, Mitrovic, Antonucci, and Rinaldi}]{kanjirangat2020sst}
Vani Kanjirangat, Sandra Mitrovic, Alessandro Antonucci, and Fabio Rinaldi. 2020.
\newblock \href {https://doi.org/10.18653/v1/2020.semeval-1.26} {{SST-BERT at SemEval-2020 Task 1: Semantic Shift Tracing by Clustering in BERT-based Embedding Spaces}}.
\newblock In \emph{Proceedings of the Fourteenth Workshop on Semantic Evaluation}, pages 214--221, Barcelona (online). International Committee for Computational Linguistics.

\bibitem[{Karjus(2023)}]{karjus2023machineassisted}
Andres Karjus. 2023.
\newblock \href {https://doi.org/doi.org/10.48550/arXiv.2309.14379} {{Machine-assisted Mixed Methods: Augmenting Humanities and Social Sciences with Artificial Intelligence}}.

\bibitem[{Kashleva et~al.(2022)Kashleva, Shein, Tukhtina, and Vydrina}]{kashleva2022black}
Kseniia Kashleva, Alexander Shein, Elizaveta Tukhtina, and Svetlana Vydrina. 2022.
\newblock \href {https://doi.org/10.18653/v1/2022.lchange-1.21} {{HSE at LSCDiscovery in Spanish: Clustering and Profiling for Lexical Semantic Change Discovery}}.
\newblock In \emph{Proceedings of the 3rd Workshop on Computational Approaches to Historical Language Change}, pages 193--197, Dublin, Ireland. Association for Computational Linguistics.

\bibitem[{Keidar et~al.(2022)Keidar, Opedal, Jin, and Sachan}]{keidar2022slangvolution}
Daphna Keidar, Andreas Opedal, Zhijing Jin, and Mrinmaya Sachan. 2022.
\newblock \href {https://doi.org/10.18653/v1/2022.acl-long.101} {{Slangvolution: A Causal Analysis of Semantic Change and Frequency Dynamics in Slang}}.
\newblock In \emph{Proceedings of the 60th Annual Meeting of the Association for Computational Linguistics (Volume 1: Long Papers)}, pages 1422--1442, Dublin, Ireland. Association for Computational Linguistics.

\bibitem[{Kocoń et~al.(2023)Kocoń, Cichecki, Kaszyca, Kochanek, Szydło, Baran, Bielaniewicz, Gruza, Janz, Kanclerz, Kocoń, Koptyra, Mieleszczenko-Kowszewicz, Miłkowski, Oleksy, Piasecki, Łukasz Radliński, Wojtasik, Woźniak, and Kazienko}]{kocon2023chatgpt}
Jan Kocoń, Igor Cichecki, Oliwier Kaszyca, Mateusz Kochanek, Dominika Szydło, Joanna Baran, Julita Bielaniewicz, Marcin Gruza, Arkadiusz Janz, Kamil Kanclerz, Anna Kocoń, Bartłomiej Koptyra, Wiktoria Mieleszczenko-Kowszewicz, Piotr Miłkowski, Marcin Oleksy, Maciej Piasecki, Łukasz Radliński, Konrad Wojtasik, Stanisław Woźniak, and Przemysław Kazienko. 2023.
\newblock \href {https://doi.org/https://doi.org/10.1016/j.inffus.2023.101861} {{ChatGPT: Jack of All Trades, Master of None}}.
\newblock \emph{Information Fusion}, 99:101861.

\bibitem[{Kudisov and Arefyev(2022)}]{kudisov2022black}
Artem Kudisov and Nikolay Arefyev. 2022.
\newblock \href {https://doi.org/10.18653/v1/2022.lchange-1.17} {{BOS at LSCDiscovery: Lexical Substitution for Interpretable Lexical Semantic Change Detection}}.
\newblock In \emph{Proceedings of the 3rd Workshop on Computational Approaches to Historical Language Change}, pages 165--172, Dublin, Ireland. Association for Computational Linguistics.

\bibitem[{Kutuzov and Giulianelli(2020)}]{kutuzov2020uio}
Andrey Kutuzov and Mario Giulianelli. 2020.
\newblock \href {https://doi.org/10.18653/v1/2020.semeval-1.14} {{UiO-UvA at SemEval-2020 Task 1: Contextualised Embeddings for Lexical Semantic Change Detection}}.
\newblock In \emph{Proceedings of the Fourteenth Workshop on Semantic Evaluation}, pages 126--134, Barcelona (online). International Committee for Computational Linguistics.

\bibitem[{Kutuzov et~al.(2018)Kutuzov, {\O}vrelid, Szymanski, and Velldal}]{kutuzov2018diachronic}
Andrey Kutuzov, Lilja {\O}vrelid, Terrence Szymanski, and Erik Velldal. 2018.
\newblock \href {https://aclanthology.org/C18-1117} {{Diachronic Word Embeddings and Semantic Shifts: a Survey}}.
\newblock In \emph{Proceedings of the 27th International Conference on Computational Linguistics}, pages 1384--1397, Santa Fe, New Mexico, USA. Association for Computational Linguistics.

\bibitem[{Kutuzov and Pivovarova(2021{\natexlab{a}})}]{kutuzov2022dwugru}
Andrey Kutuzov and Lidia Pivovarova. 2021{\natexlab{a}}.
\newblock \href {https://github.com/akutuzov/rushifteval_public} {{RuShiftEval}}.

\bibitem[{Kutuzov and Pivovarova(2021{\natexlab{b}})}]{kutuzov2021rushifteval}
Andrey Kutuzov and Lidia Pivovarova. 2021{\natexlab{b}}.
\newblock \href {https://www.dialog-21.ru/media/5536/pivovarovalpluskutuzova151.pdf} {{RuShiftEval: A Shared Task on Semantic Shift Detection for Russian}}.
\newblock In \emph{Proceedings of the Conference on Computational Linguistics and Intellectual Technologies (Dialogue)}, 20, (online). RSUH.

\bibitem[{Kutuzov and Pivovarova(2021{\natexlab{c}})}]{kutuzov2021three}
Andrey Kutuzov and Lidia Pivovarova. 2021{\natexlab{c}}.
\newblock \href {https://doi.org/10.18653/v1/2021.lchange-1.2} {{Three-part Diachronic Semantic Change Dataset for Russian}}.
\newblock In \emph{Proceedings of the 2nd International Workshop on Computational Approaches to Historical Language Change 2021}, pages 7--13, Online. Association for Computational Linguistics.

\bibitem[{Kutuzov et~al.(2022{\natexlab{a}})Kutuzov, Touileb, M{\ae}hlum, Enstad, and Wittemann}]{kutuzov2022nordiachange}
Andrey Kutuzov, Samia Touileb, Petter M{\ae}hlum, Tita Enstad, and Alexandra Wittemann. 2022{\natexlab{a}}.
\newblock \href {https://aclanthology.org/2022.lrec-1.274} {{NorDiaChange: Diachronic Semantic Change Dataset for Norwegian}}.
\newblock In \emph{Proceedings of the Thirteenth Language Resources and Evaluation Conference}, pages 2563--2572, Marseille, France. European Language Resources Association.

\bibitem[{Kutuzov et~al.(2021)Kutuzov, Touileb, Mæhlum, Ranveig~Enstad, and Wittemann}]{kutuzov2022dwugno}
Andrey Kutuzov, Samia Touileb, Petter Mæhlum, Tita Ranveig~Enstad, and Alexandra Wittemann. 2021.
\newblock \href {https://github.com/ltgoslo/nor_dia_change} {{NorDiaChange}}.

\bibitem[{Kutuzov et~al.(2022{\natexlab{b}})Kutuzov, Velldal, and {\O}vrelid}]{kutuzov2022contextualized}
Andrey Kutuzov, Erik Velldal, and Lilja {\O}vrelid. 2022{\natexlab{b}}.
\newblock \href {https://doi.org/https://doi.org/10.3384/nejlt.2000-1533.2022.3478} {{Contextualized Embeddings for Semantic Change Detection: Lessons Learned}}.
\newblock In \emph{Northern European Journal of Language Technology, Volume 8}, Copenhagen, Denmark. Northern European Association of Language Technology.

\bibitem[{Laicher et~al.(2021)Laicher, Kurtyigit, Schlechtweg, Kuhn, and Schulte~im Walde}]{laicher2021explaining}
Severin Laicher, Sinan Kurtyigit, Dominik Schlechtweg, Jonas Kuhn, and Sabine Schulte~im Walde. 2021.
\newblock \href {https://doi.org/10.18653/v1/2021.eacl-srw.25} {{Explaining and Improving BERT Performance on Lexical Semantic Change Detection}}.
\newblock In \emph{Proceedings of the 16th Conference of the European Chapter of the Association for Computational Linguistics: Student Research Workshop}, pages 192--202, Online. Association for Computational Linguistics.

\bibitem[{Laskar et~al.(2023)Laskar, Bari, Rahman, Bhuiyan, Joty, and Huang}]{laskar2023systematic}
Md~Tahmid~Rahman Laskar, M~Saiful Bari, Mizanur Rahman, Md~Amran~Hossen Bhuiyan, Shafiq Joty, and Jimmy Huang. 2023.
\newblock \href {https://doi.org/10.18653/v1/2023.findings-acl.29} {{A Systematic Study and Comprehensive Evaluation of ChatGPT on Benchmark Datasets}}.
\newblock In \emph{Findings of the Association for Computational Linguistics: ACL 2023}, pages 431--469, Toronto, Canada. Association for Computational Linguistics.

\bibitem[{Lendvai and Wick(2022)}]{lendvai2022finetuning}
Piroska Lendvai and Claudia Wick. 2022.
\newblock \href {https://aclanthology.org/2022.cogalex-1.5} {{Finetuning Latin BERT for Word Sense Disambiguation on the Thesaurus Linguae Latinae}}.
\newblock In \emph{Proceedings of the Workshop on Cognitive Aspects of the Lexicon}, pages 37--41, Taipei, Taiwan. Association for Computational Linguistics.

\bibitem[{Li et~al.(2023)Li, Wang, Zhang, Zhu, Hou, Lian, Luo, Yang, and Xie}]{li2023large}
Cheng Li, Jindong Wang, Yixuan Zhang, Kaijie Zhu, Wenxin Hou, Jianxun Lian, Fang Luo, Qiang Yang, and Xing Xie. 2023.
\newblock \href {https://doi.org/doi.org/10.48550/arXiv.2307.11760} {{Large Language Models Understand and Can be Enhanced by Emotional Stimuli}}.

\bibitem[{Liang and Shi(2023)}]{liang2023named}
Meng Liang and Yao Shi. 2023.
\newblock \href {https://doi.org/10.1109/ICNLP58431.2023.00041} {{Named Entity Recognition Method Based on BERT-whitening and Dynamic Fusion Model}}.
\newblock In \emph{2023 5th International Conference on Natural Language Processing (ICNLP)}, pages 191--197.

\bibitem[{Loureiro et~al.(2022)Loureiro, D{'}Souza, Muhajab, White, Wong, Espinosa-Anke, Neves, Barbieri, and Camacho-Collados}]{loureiro2022tempowic}
Daniel Loureiro, Aminette D{'}Souza, Areej~Nasser Muhajab, Isabella~A. White, Gabriel Wong, Luis Espinosa-Anke, Leonardo Neves, Francesco Barbieri, and Jose Camacho-Collados. 2022.
\newblock \href {https://aclanthology.org/2022.coling-1.296} {{TempoWiC: An Evaluation Benchmark for Detecting Meaning Shift in Social Media}}.
\newblock In \emph{Proceedings of the 29th International Conference on Computational Linguistics}, pages 3353--3359, Gyeongju, Republic of Korea. International Committee on Computational Linguistics.

\bibitem[{Ma et~al.(2019)Ma, Wang, Ng, Nallapati, and Xiang}]{ma2019universal}
Xiaofei Ma, Zhiguo Wang, Patrick Ng, Ramesh Nallapati, and Bing Xiang. 2019.
\newblock \href {https://doi.org/doi.org/10.48550/arXiv.1910.07973} {{Universal Text Representation from BERT: An Empirical Study}}.

\bibitem[{Manandhar et~al.(2010)Manandhar, Klapaftis, Dligach, and Pradhan}]{manandhar2010semeval}
Suresh Manandhar, Ioannis Klapaftis, Dmitriy Dligach, and Sameer Pradhan. 2010.
\newblock \href {https://aclanthology.org/S10-1011} {{SemEval-2010 Task 14: Word Sense Induction {\&} Disambiguation}}.
\newblock In \emph{Proceedings of the 5th International Workshop on Semantic Evaluation}, pages 63--68, Uppsala, Sweden. Association for Computational Linguistics.

\bibitem[{Manning(2009)}]{manning2009introduction}
Christopher~D Manning. 2009.
\newblock \href {https://nlp.stanford.edu/IR-book/information-retrieval-book.html} {\emph{{An Introduction to Information Retrieval}}}.
\newblock Cambridge university press.

\bibitem[{Martinc et~al.(2020{\natexlab{a}})Martinc, Kralj~Novak, and Pollak}]{martinc2020leveraging}
Matej Martinc, Petra Kralj~Novak, and Senja Pollak. 2020{\natexlab{a}}.
\newblock \href {https://aclanthology.org/2020.lrec-1.592} {{Leveraging Contextual Embeddings for Detecting Diachronic Semantic Shift}}.
\newblock In \emph{Proceedings of the Twelfth Language Resources and Evaluation Conference}, pages 4811--4819, Marseille, France. European Language Resources Association.

\bibitem[{Martinc et~al.(2020{\natexlab{b}})Martinc, Montariol, Zosa, and Pivovarova}]{martinc2020capturing}
Matej Martinc, Syrielle Montariol, Elaine Zosa, and Lidia Pivovarova. 2020{\natexlab{b}}.
\newblock \href {https://doi.org/10.1145/3366424.3382186} {{Capturing Evolution in Word Usage: Just Add More Clusters?}}
\newblock In \emph{Companion Proceedings of the Web Conference 2020}, WWW '20, page 343–349, Taipei, Taiwan. Association for Computing Machinery.

\bibitem[{Martinc et~al.(2020{\natexlab{c}})Martinc, Montariol, Zosa, and Pivovarova}]{martinc2020discovery}
Matej Martinc, Syrielle Montariol, Elaine Zosa, and Lidia Pivovarova. 2020{\natexlab{c}}.
\newblock \href {https://doi.org/10.18653/v1/2020.semeval-1.6} {{Discovery Team at SemEval-2020 Task 1: Context-sensitive Embeddings Not Always Better than Static for Semantic Change Detection}}.
\newblock In \emph{Proceedings of the Fourteenth Workshop on Semantic Evaluation}, pages 67--73, Barcelona (online). International Committee for Computational Linguistics.

\bibitem[{McGillivray et~al.(2021)McGillivray, Schlechtweg, Dubossarsky, Tahmasebi, and Hengchen}]{barbara2021dwugla}
Barbara McGillivray, Dominik Schlechtweg, Haim Dubossarsky, Nina Tahmasebi, and Simon Hengchen. 2021.
\newblock \href {https://doi.org/10.5281/zenodo.5255228} {{DWUG LA: Diachronic Word Usage Graphs for Latin}}.

\bibitem[{Montanelli and Periti(2023)}]{montanelli2023survey}
Stefano Montanelli and Francesco Periti. 2023.
\newblock \href {https://doi.org/https://doi.org/10.48550/arXiv.2304.01666} {{A Survey on Contextualised Semantic Shift Detection}}.

\bibitem[{Montariol et~al.(2021)Montariol, Martinc, and Pivovarova}]{montariol2021scalable}
Syrielle Montariol, Matej Martinc, and Lidia Pivovarova. 2021.
\newblock \href {https://doi.org/10.18653/v1/2021.naacl-main.369} {{Scalable and Interpretable Semantic Change Detection}}.
\newblock In \emph{Proceedings of the 2021 Conference of the North American Chapter of the Association for Computational Linguistics: Human Language Technologies}, pages 4642--4652, Online. Association for Computational Linguistics.

\bibitem[{Periti and Dubossarsky(2023)}]{periti2023time}
Francesco Periti and Haim Dubossarsky. 2023.
\newblock \href {https://ceur-ws.org/Vol-3473/paper47.pdf} {{The Time-Embedding Travelers@WiC-ITA}}.
\newblock In \emph{Proceedings of the Eighth Evaluation Campaign of Natural Language Processing and Speech Tools for Italian. Final Workshop (EVALITA 2023)}, Parma, Italy. CEUR.org.

\bibitem[{Periti et~al.(2024)Periti, Dubossarsky, and Tahmasebi}]{periti2024chatgpt}
Francesco Periti, Haim Dubossarsky, and Nina Tahmasebi. 2024.
\newblock \href {http://arxiv.org/abs/2401.14040} {{(Chat)GPT v BERT: Dawn of Justice for Semantic Change Detection}}.

\bibitem[{Periti et~al.(2022)Periti, Ferrara, Montanelli, and Ruskov}]{periti2022what}
Francesco Periti, Alfio Ferrara, Stefano Montanelli, and Martin Ruskov. 2022.
\newblock \href {https://doi.org/10.18653/v1/2022.lchange-1.4} {{What is Done is Done: an Incremental Approach to Semantic Shift Detection}}.
\newblock In \emph{Proceedings of the 3rd Workshop on Computational Approaches to Historical Language Change}, pages 33--43, Dublin, Ireland. Association for Computational Linguistics.

\bibitem[{Periti et~al.(2023)Periti, Picascia, Montanelli, Ferrara, and Tahmasebi}]{periti2023word}
Francesco Periti, Sergio Picascia, Stefano Montanelli, Alfio Ferrara, and Nina Tahmasebi. 2023.
\newblock \href {https://doi.org/10.36227/techrxiv.24210915.v1} {{Studying Word Meaning Evolution through Incremental Semantic Shift Detection: A Case Study of Italian Parliamentary Speeches}}.

\bibitem[{Pilehvar and Camacho-Collados(2019)}]{pilehvar2019wic}
Mohammad~Taher Pilehvar and Jose Camacho-Collados. 2019.
\newblock \href {https://doi.org/10.18653/v1/N19-1128} {{WiC: the Word-in-Context Dataset for Evaluating Context-Sensitive Meaning Representations}}.
\newblock In \emph{Proceedings of the 2019 Conference of the North {A}merican Chapter of the Association for Computational Linguistics: Human Language Technologies, Volume 1 (Long and Short Papers)}, pages 1267--1273, Minneapolis, Minnesota. Association for Computational Linguistics.

\bibitem[{P{\"o}msl and Lyapin(2020)}]{pomsl2020circe}
Martin P{\"o}msl and Roman Lyapin. 2020.
\newblock \href {https://doi.org/10.18653/v1/2020.semeval-1.21} {{CIRCE at SemEval-2020 Task 1: Ensembling Context-Free and Context-Dependent Word Representations}}.
\newblock In \emph{Proceedings of the Fourteenth Workshop on Semantic Evaluation}, pages 180--186, Barcelona (online). International Committee for Computational Linguistics.

\bibitem[{Rachinskiy and Arefyev(2022)}]{rachinskiy2022black}
Maxim Rachinskiy and Nikolay Arefyev. 2022.
\newblock \href {https://doi.org/10.18653/v1/2022.lchange-1.22} {{GlossReader at LSCDiscovery: Train to Select a Proper Gloss in English -- Discover Lexical Semantic Change in Spanish}}.
\newblock In \emph{Proceedings of the 3rd Workshop on Computational Approaches to Historical Language Change}, pages 198--203, Dublin, Ireland. Association for Computational Linguistics.

\bibitem[{Raganato et~al.(2020)Raganato, Pasini, Camacho-Collados, and Pilehvar}]{raganato2020xl}
Alessandro Raganato, Tommaso Pasini, Jose Camacho-Collados, and Mohammad~Taher Pilehvar. 2020.
\newblock \href {https://doi.org/10.18653/v1/2020.emnlp-main.584} {{XL-WiC: A Multilingual Benchmark for Evaluating Semantic Contextualization}}.
\newblock In \emph{Proceedings of the 2020 Conference on Empirical Methods in Natural Language Processing (EMNLP)}, pages 7193--7206, Online. Association for Computational Linguistics.

\bibitem[{Reif et~al.(2019)Reif, Yuan, Wattenberg, Viegas, Coenen, Pearce, and Kim}]{coenen2019visualizing}
Emily Reif, Ann Yuan, Martin Wattenberg, Fernanda~B Viegas, Andy Coenen, Adam Pearce, and Been Kim. 2019.
\newblock \href {https://proceedings.neurips.cc/paper_files/paper/2019/file/159c1ffe5b61b41b3c4d8f4c2150f6c4-Paper.pdf} {{Visualizing and Measuring the Geometry of BERT}}.
\newblock In \emph{Advances in Neural Information Processing Systems}, volume~32. Curran Associates, Inc.

\bibitem[{Riemenschneider and Frank(2023)}]{riemenschneider2023exploring}
Frederick Riemenschneider and Anette Frank. 2023.
\newblock \href {https://doi.org/10.18653/v1/2023.acl-long.846} {{Exploring Large Language Models for Classical Philology}}.
\newblock In \emph{Proceedings of the 61st Annual Meeting of the Association for Computational Linguistics (Volume 1: Long Papers)}, pages 15181--15199, Toronto, Canada. Association for Computational Linguistics.

\bibitem[{Rodina et~al.(2021)Rodina, Trofimova, Kutuzov, and Artemova}]{rodina2021elmo}
Julia Rodina, Yuliya Trofimova, Andrey Kutuzov, and Ekaterina Artemova. 2021.
\newblock {ELMo and BERT in Semantic Change Detection for Russian}.
\newblock In \emph{Analysis of Images, Social Networks and Texts}, pages 175--186, Cham. Springer International Publishing.

\bibitem[{Rosin et~al.(2022)Rosin, Guy, and Radinsky}]{rosin2022time}
Guy~D. Rosin, Ido Guy, and Kira Radinsky. 2022.
\newblock \href {https://doi.org/10.1145/3488560.3498529} {{Time Masking for Temporal Language Models}}.
\newblock In \emph{Proceedings of the Fifteenth ACM International Conference on Web Search and Data Mining}, WSDM '22, page 833–841, Virtual Event, AZ, USA. Association for Computing Machinery.

\bibitem[{Schlechtweg(2023)}]{schlechtweg2023human}
Dominik Schlechtweg. 2023.
\newblock \href {https://doi.org/http://dx.doi.org/10.18419/opus-12833} {\emph{{Human and Computational Measurement of Lexical Semantic Change}}}.
\newblock Ph.D. thesis, University of Stuttgart.

\bibitem[{Schlechtweg et~al.(2022{\natexlab{a}})Schlechtweg, Dubossarsky, Hengchen, McGillivray, and Tahmasebi}]{schlechtweg2022dwugen}
Dominik Schlechtweg, Haim Dubossarsky, Simon Hengchen, Barbara McGillivray, and Nina Tahmasebi. 2022{\natexlab{a}}.
\newblock \href {https://doi.org/10.5281/zenodo.7387261} {{DWUG EN: Diachronic Word Usage Graphs for English}}.

\bibitem[{Schlechtweg et~al.(2020)Schlechtweg, McGillivray, Hengchen, Dubossarsky, and Tahmasebi}]{schlechtweg2020semeval}
Dominik Schlechtweg, Barbara McGillivray, Simon Hengchen, Haim Dubossarsky, and Nina Tahmasebi. 2020.
\newblock \href {https://doi.org/10.18653/v1/2020.semeval-1.1} {{SemEval-2020 Task 1: Unsupervised Lexical Semantic Change Detection}}.
\newblock In \emph{Proceedings of the Fourteenth Workshop on Semantic Evaluation}, pages 1--23, Barcelona (online). International Committee for Computational Linguistics.

\bibitem[{Schlechtweg et~al.(2022{\natexlab{b}})Schlechtweg, McGillivray, Hengchen, Dubossarsky, and Tahmasebi}]{schlechtweg2022dwugde}
Dominik Schlechtweg, Barbara McGillivray, Simon Hengchen, Haim Dubossarsky, and Nina Tahmasebi. 2022{\natexlab{b}}.
\newblock \href {https://doi.org/10.5281/zenodo.7441645} {{DWUG DE: Diachronic Word Usage Graphs for German}}.

\bibitem[{Schlechtweg and Schulte~im Walde(2020)}]{schlechtweg2020simulating}
Dominik Schlechtweg and Sabine Schulte~im Walde. 2020.
\newblock \href {https://brussels.evolang.org/proceedings/papers/EvoLang13_paper_9.pdf} {{Simulating Lexical Semantic Change from Sense-Annotated Data}}.
\newblock In \emph{Proceedings of the 13th International Conference on the Evolution of Language (EvoLang13)}, Brussels, Belgium.

\bibitem[{Schlechtweg et~al.(2018)Schlechtweg, Schulte~im Walde, and Eckmann}]{schlechtweg2018diachronic}
Dominik Schlechtweg, Sabine Schulte~im Walde, and Stefanie Eckmann. 2018.
\newblock \href {https://doi.org/10.18653/v1/N18-2027} {{Diachronic Usage Relatedness (DURel): A Framework for the Annotation of Lexical Semantic Change}}.
\newblock In \emph{Proceedings of the 2018 Conference of the North {A}merican Chapter of the Association for Computational Linguistics: Human Language Technologies, Volume 2 (Short Papers)}, pages 169--174, New Orleans, Louisiana. Association for Computational Linguistics.

\bibitem[{Schlechtweg et~al.(2021)Schlechtweg, Tahmasebi, Hengchen, Dubossarsky, and McGillivray}]{schlechtweg2021dwug}
Dominik Schlechtweg, Nina Tahmasebi, Simon Hengchen, Haim Dubossarsky, and Barbara McGillivray. 2021.
\newblock \href {https://doi.org/10.18653/v1/2021.emnlp-main.567} {{DWUG: A large Resource of Diachronic Word Usage Graphs in Four Languages}}.
\newblock In \emph{Proceedings of the 2021 Conference on Empirical Methods in Natural Language Processing}, pages 7079--7091, Online and Punta Cana, Dominican Republic. Association for Computational Linguistics.

\bibitem[{Schlechtweg et~al.(2023)Schlechtweg, Virk, Sander, Sköldberg, Linke, Zhang, Tahmasebi, Kuhn, and im~Walde}]{schlechtweg2023durel}
Dominik Schlechtweg, Shafqat~Mumtaz Virk, Pauline Sander, Emma Sköldberg, Lukas~Theuer Linke, Tuo Zhang, Nina Tahmasebi, Jonas Kuhn, and Sabine~Schulte im~Walde. 2023.
\newblock \href {https://doi.org/doi.org/10.48550/arXiv.2311.12664} {{The DURel Annotation Tool: Human and Computational Measurement of Semantic Proximity, Sense Clusters and Semantic Change}}.

\bibitem[{Shoemark et~al.(2019)Shoemark, Liza, Nguyen, Hale, and McGillivray}]{shoemark2019room}
Philippa Shoemark, Farhana~Ferdousi Liza, Dong Nguyen, Scott Hale, and Barbara McGillivray. 2019.
\newblock \href {https://doi.org/10.18653/v1/D19-1007} {{Room to Glo: A Systematic Comparison of Semantic Change Detection Approaches with Word Embeddings}}.
\newblock In \emph{Proceedings of the 2019 Conference on Empirical Methods in Natural Language Processing and the 9th International Joint Conference on Natural Language Processing (EMNLP-IJCNLP)}, pages 66--76, Hong Kong, China. Association for Computational Linguistics.

\bibitem[{Tahmasebi et~al.(2021)Tahmasebi, Borin, and Jatowt}]{tahmasebi2021survey}
Nina Tahmasebi, Lars Borin, and Adam Jatowt. 2021.
\newblock \href {https://doi.org/10.5281/zenodo.5040302} {\emph{{Survey of Computational Approaches to Lexical Semantic Change Detection}}}, pages 1--91. Language Science Press, Berlin.

\bibitem[{Tahmasebi et~al.(2022)Tahmasebi, Hengchen, Schlechtweg, McGillivray, and Dubossarsky}]{tahmasebi2022dwugsv}
Nina Tahmasebi, Simon Hengchen, Dominik Schlechtweg, Barbara McGillivray, and Haim Dubossarsky. 2022.
\newblock \href {https://doi.org/10.5281/zenodo.7389506} {{DWUG SV: Diachronic Word Usage Graphs for Swedish}}.

\bibitem[{Tang et~al.(2023)Tang, Zhou, Aida, Sen, and Bollegala}]{tang2023word}
Xiaohang Tang, Yi~Zhou, Taichi Aida, Procheta Sen, and Danushka Bollegala. 2023.
\newblock \href {https://doi.org/10.18653/v1/2023.findings-emnlp.231} {{Can Word Sense Distribution Detect Semantic Changes of Words?}}
\newblock In \emph{Findings of the Association for Computational Linguistics: EMNLP 2023}, pages 3575--3590, Singapore. Association for Computational Linguistics.

\bibitem[{Tang(2018)}]{tang2018state}
Xuri Tang. 2018.
\newblock \href {https://doi.org/10.1017/S1351324918000220} {{A State-of-the-art of Semantic Change Computation}}.
\newblock \emph{Natural Language Engineering}, 24(5):649–676.

\bibitem[{Wolf et~al.(2020)Wolf, Debut, Sanh, Chaumond, Delangue, Moi, Cistac, Rault, Louf, Funtowicz, Davison, Shleifer, von Platen, Ma, Jernite, Plu, Xu, Le~Scao, Gugger, Drame, Lhoest, and Rush}]{wolf2020transformers}
Thomas Wolf, Lysandre Debut, Victor Sanh, Julien Chaumond, Clement Delangue, Anthony Moi, Pierric Cistac, Tim Rault, Remi Louf, Morgan Funtowicz, Joe Davison, Sam Shleifer, Patrick von Platen, Clara Ma, Yacine Jernite, Julien Plu, Canwen Xu, Teven Le~Scao, Sylvain Gugger, Mariama Drame, Quentin Lhoest, and Alexander Rush. 2020.
\newblock \href {https://doi.org/10.18653/v1/2020.emnlp-demos.6} {{Transformers: State-of-the-Art Natural Language Processing}}.
\newblock In \emph{Proceedings of the 2020 Conference on Empirical Methods in Natural Language Processing: System Demonstrations}, pages 38--45, Online. Association for Computational Linguistics.

\bibitem[{Zamora-Reina et~al.(2022{\natexlab{a}})Zamora-Reina, Bravo-Marquez, and Schlechtweg}]{zamora2022dwuges}
Frank~D. Zamora-Reina, Felipe Bravo-Marquez, and Dominik Schlechtweg. 2022{\natexlab{a}}.
\newblock \href {https://doi.org/10.5281/zenodo.6433667} {Dwug es: Diachronic word usage graphs for spanish}.

\bibitem[{Zamora-Reina et~al.(2022{\natexlab{b}})Zamora-Reina, Bravo-Marquez, and Schlechtweg}]{zamora2022black}
Frank~D. Zamora-Reina, Felipe Bravo-Marquez, and Dominik Schlechtweg. 2022{\natexlab{b}}.
\newblock \href {https://doi.org/10.18653/v1/2022.lchange-1.16} {{LSCDiscovery: A Shared Task on Semantic Change Discovery and Detection in Spanish}}.
\newblock In \emph{Proceedings of the 3rd Workshop on Computational Approaches to Historical Language Change}, pages 149--164, Dublin, Ireland. Association for Computational Linguistics.

\bibitem[{Zhou and Li(2020)}]{zhou2020temporalteller}
Jinan Zhou and Jiaxin Li. 2020.
\newblock \href {https://doi.org/10.18653/v1/2020.semeval-1.27} {{TemporalTeller at SemEval-2020 Task 1: Unsupervised Lexical Semantic Change Detection with Temporal Referencing}}.
\newblock In \emph{Proceedings of the Fourteenth Workshop on Semantic Evaluation}, pages 222--231, Barcelona (online). International Committee for Computational Linguistics.

\bibitem[{Zhou et~al.(2023)Zhou, Tahmasebi, and Dubossarsky}]{zhou2023finer}
Wei Zhou, Nina Tahmasebi, and Haim Dubossarsky. 2023.
\newblock \href {https://aclanthology.org/2023.nodalida-1.52} {{The Finer They Get: Combining Fine-Tuned Models For Better Semantic Change Detection}}.
\newblock In \emph{Proceedings of the 24th Nordic Conference on Computational Linguistics (NoDaLiDa)}, pages 518--528, T{\'o}rshavn, Faroe Islands. University of Tartu Library.

\end{thebibliography}
\clearpage
\section*{Appendix}
\appendix
\section{Semantic proximity}
As an example, consider the following word usage pair $\langle w, c_1, c_2 \rangle$ extracted by the English benchmark for the word $w=$\textit{plane}.

\begin{itemize}\label{ex:wic}
    \item $c_1$: But we are most familiar with the exhibitions of gravity in bodies descending inclined \textbf{planes}, as in the avalanche and the cataract.
    \item $c_2$: Over the next several years, he said, the Coast Guard will get 60 more people, two new 270-foot vessels and al twin-engine \textbf{planes}.
\end{itemize}

Following the DURel relatedness scale (see Table~\ref{durel:scale}), the pair is annotated with an average judgment of \textbf{1} by human annotators.

\begin{table}[!ht]
\centering
\resizebox{0.60\columnwidth}{!}{%
\begin{tabular}{lcl}
\multirow{4}{*}{$\Bigg\uparrow$} & 4: & Identical         \\
                                 & 3: & Closely related   \\
                                 & 2: & Distantly related \\
                                 & 1: & Unrelated        
\end{tabular}}
\caption{The DURel relatedness scale used in~\newcite{schlechtweg2023durel,schlechtweg2023human,schlechtweg2021dwug,schlechtweg2020semeval,schlechtweg2020simulating,schlechtweg2018diachronic}}
\label{durel:scale}
\end{table}

\section{State-of-the-art for Graded Change Detection}
In Table~\ref{tab:sota}, we report the current top scores for GCD in the state-of-the-art with a reference to the paper from where the result is taken. Notably, we report results for different benchmarks using four different approaches evaluated in this paper. However, some benchmarks might have been assessed using other approaches that are excluded from this table. 

\section{Graded Change Detection across layers}\label{app:layers}
In Table~\ref{tab:all-layers}, we report correlation scores for GCD across benchmarks. Specifically, we report results for BERT, mBERT, and XLM-R (separated by slash, i.e. ``/'') by utilizing all layers of the models (1-12), \textit{individually}.

\begin{figure*}
    \centering
    \includegraphics[width=\textwidth]{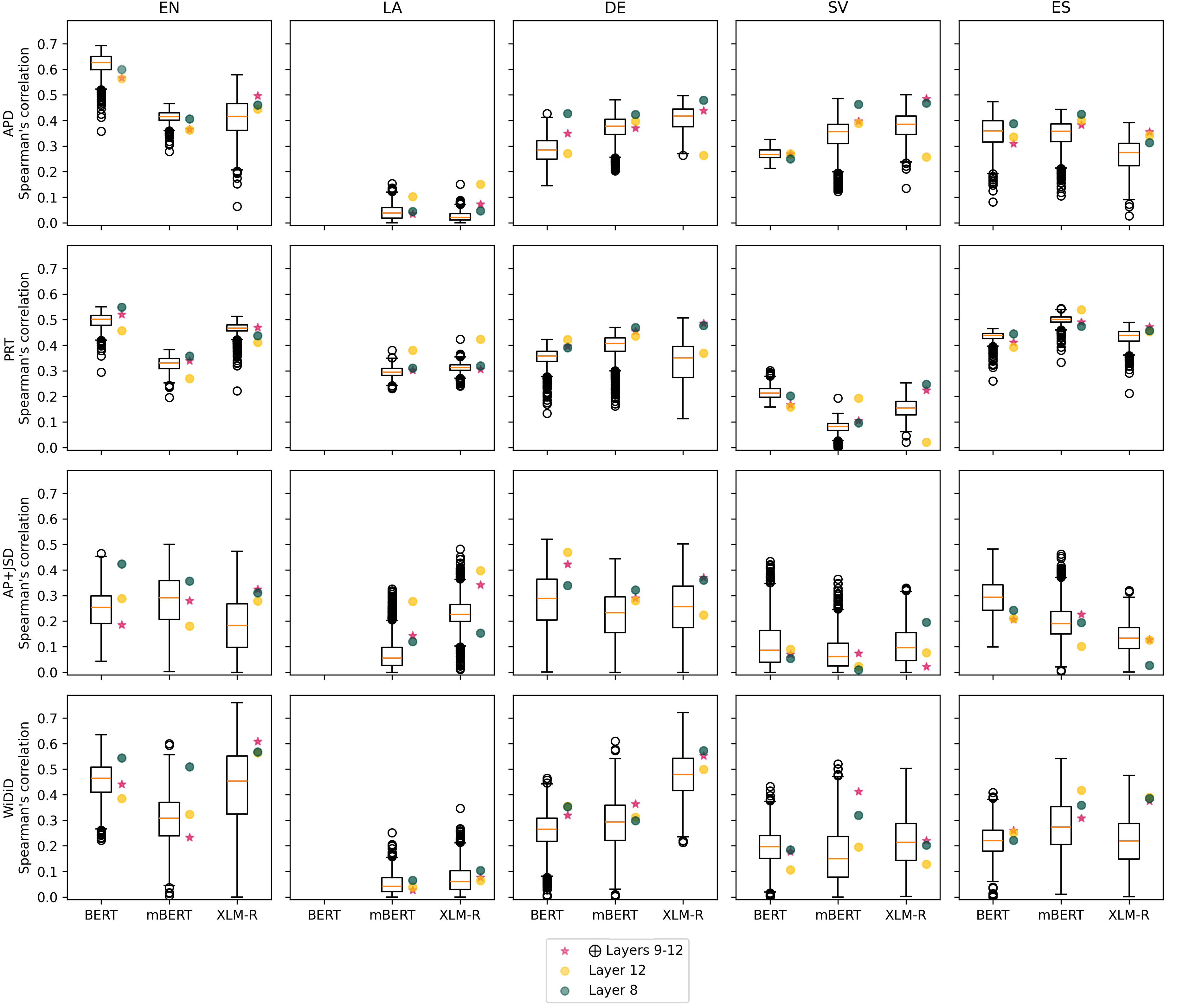}
    \caption{\textbf{Score distribution for GCD} obtained by using all possible layer combinations of length 2 (e.g., Layer 1 and 2), length 3 (e.g., Layer 10, 11, 12), and length 4 (e.g., Layer 1, 10, 11, 12) for BERT, mBERT, and XLM-R. The y-axis represents the Spearman correlation. We highlight the performance for GCD obtained using Layer 8, Layer 12, and the sum of the last 4 layers (i.e., $\bigoplus$ 9-12).}
    \label{fig:layers}
\end{figure*}

\begin{figure*}
    \centering
    \includegraphics[width=\textwidth]{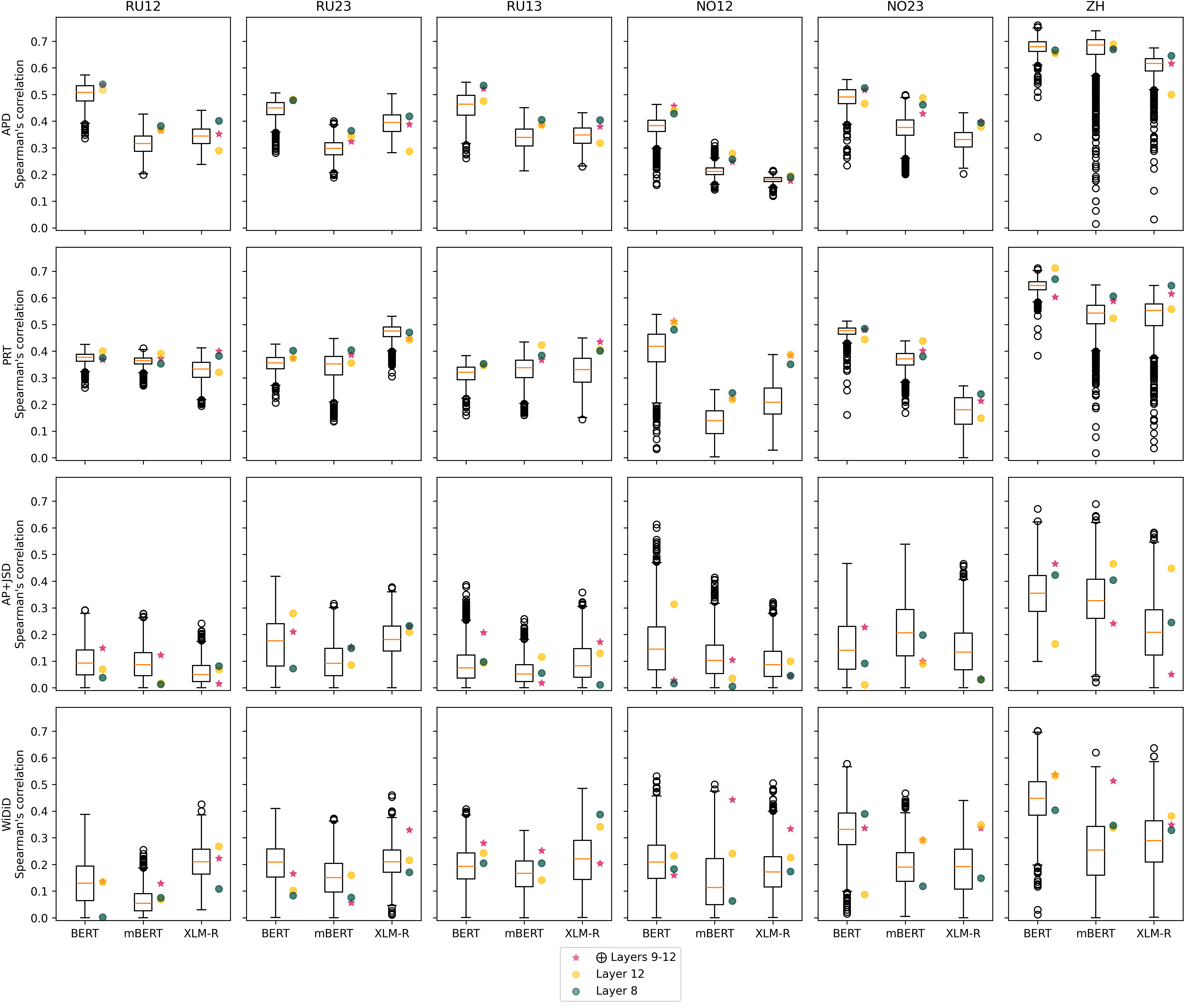}
    \caption{\textbf{Score distribution for GCD} obtained by using all possible layer combinations of length 2 (e.g., Layer 1 and 2), length 3 (e.g., Layer 10, 11, 12), and length 4 (e.g., Layer 1, 10, 11, 12) for BERT, mBERT, and XLM-R. The y-axis represents the Spearman correlation. We highlight the performance for GCD obtained using Layer 8, Layer 12, and the sum of the last 4 layers (i.e., $\bigoplus$ 9-12).}
    \label{fig:layers1}
\end{figure*}

In Figure~\ref{fig:layers} and~\ref{fig:layers1}, we report correlation scores distribution for GCD obtained by using all possible layer \textit{combinations} of length 2 (e.g., Layer 1 and 2), length 3 (e.g., Layer 10, 11, 12), and length 4 (e.g., Layer 1, 10, 11, 12) for BERT, mBERT, and XLM-R. 

For the sake of comparison, we report in Table~\ref{tab:best-vs-last4} the overall top score for GCD obtained using BERT, mBERT, and XLM-R. Specifically, we present results for the optimal combination and the outcome obtained by summing the last four layers, separated by a slash. Additionally, we include the standard result obtained using the last layer individually.

\section{Benchmarks}
In Table~\ref{tab:benchmarks}, we report the benchmarks used in this work. Specifically, for each benchmark, we report time periods, diachronic corpus composition, number of targets, and benchmark versions.

\section{BERT, mBERT, XLM-R, XL-LEXEME}
In Table~\ref{tab:models-evaluated}, we report the BERT, mBERT, XLM-R, and XL-LEXEME models employed in our evaluation. All the models are base versions with 12 encoder layers and can be accessed on \url{huggingface.co}.

\section{BERT, mBERT, XLM-R, XL-LEXEME}
In Table~\ref{tab:models-evaluated}, we report the BERT, mBERT, XLM-R, and XL-LEXEME models employed in our evaluation. All the models are base versions with 12 encoder layers and can be accessed on \url{huggingface.co}.

\newpage

\section{GPT-4 evaluation}
We evaluate GPT-4 as computational annotator by relying on computational proximity judgments gathered through the following method.

\paragraph{Model initialization.}
We initialized the model with the following prompt (guideline):\\\\
{\small 
\tt
Determine whether an input word has the same meaning in the two input sentences. Answer with 'Same', 'Related', 'Linked', or 'Distinct'. This is very important to my career.
}
\\\\
\noindent Notably, we combine and refine two different prompts used in previous works. We drew inspiration from the prompt utilized by~\citet{karjus2023machineassisted} to assess GPT-4 in addressing the Graded Change Detection task. Additionally, we drew inspiration from the prompt utilized by~\citet{li2023large}, called \textit{EmotionPrompt}, which combines the original prompt with emotional stimuli to enhance the performance of Large Language Models.

\paragraph{Model template.}
For each word usage pair, we used the following prompt:\\\\ 
{\small
\tt
Determine whether \textit{[Target word]} has the same meaning in the following sentences. Do they refer to roughly the Same, different but closely Related, distant/figuratively Linked or unrelated Distinct word meanings?\\
\textbf{Sentence 1:} \textit{[Context 1]}\\
\textbf{Sentence 2:} \textit{[Context 2]}\\
}
\\
\noindent Notably, drawing inspiration from the OpenAI documentation\footnote{\url{platform.openai.com/docs/guides/prompt-engineering}} and the prompts utilized in previous work for the Word-in-Context task~\cite{kocon2023chatgpt,laskar2023systematic}, we structured our prompt in a format that facilitates parsing and comprehension. For each usage pair $\langle w, c_1, c_2\rangle$ of a word $w$, we substitute [Target word] with the actual target $w$ and [Context 1] and [Context 2] with $c_1$ and $c_2$, respectively.

We prompt GPT-4 without providing any message history. This means that, for each usage pair $\langle w, c_1, c_2\rangle$, we re-initialize the model with the initial prompt (guideline) and subsequently prompt the model to gather a semantic proximity judgment for the pair $\langle w, c_1, c_2\rangle$. This approach ensures that the model relies solely on its pre-trained knowledge, preventing potential biases stemming from previously prompted pairs.

\begin{landscape}
\begin{table}[!t]
\centering
\resizebox{1.5\textwidth}{!}{%
\begin{tabular}{ccccccccccccccc}
\multicolumn{1}{l}{} &  &  & \textbf{EN} & \textbf{LA} & \textbf{DE} & \textbf{SV} & \textbf{ES} & \multicolumn{3}{c}{\textbf{RU}} & \multicolumn{2}{c}{\textbf{NO}} & \textbf{ZH} & \textbf{Avg$_w$}  \\ \cline{4-15} 
\multicolumn{1}{l}{} &  & \multicolumn{1}{c|}{} & \multicolumn{1}{c|}{\textit{$C_1 - C_2$}} & \multicolumn{1}{c|}{\textit{$C_1 - C_2$}} & \multicolumn{1}{c|}{\textit{$C_1 - C_2$}} & \multicolumn{1}{c|}{\textit{$C_1 - C_2$}} & \multicolumn{1}{c|}{\textit{$C_1 - C_2$}} & \textit{$C_1 - C_2$} & \textit{$C_2 - C_3$} & \multicolumn{1}{c|}{\textit{$C_1 - C_3$}} & \textit{$C_1 - C_2$} & \multicolumn{1}{c|}{\textit{$C_2 - C_3$}} & \multicolumn{1}{c|}{\textit{$C_1 - C_2$}} & \multicolumn{1}{c|}{\textit{$C_i - C_j$}}\\ \cline{2-15} 
\multicolumn{1}{c|}{\multirow{2}{*}{\rotatebox{90}{\textbf{form-based}} \vspace{-150pt}}} & \multicolumn{1}{c|}{APD} & \multicolumn{1}{c|}{\begin{tabular}[c]{@{}c@{}}1\\ 2\\ 3\\ 4\\ 5\\ 6\\ 7\\ 8\\ 9\\ 10\\ 11\\ 12\end{tabular}} & \multicolumn{1}{c|}{\begin{tabular}[c]{@{}c@{}}.358 / .278 / .064\\ .464 / .346 / .229\\ .574 / .389 / .314\\ .628 / .410 / .400\\ \textbf{.684} / .412 / .452\\ .667 / .395 / .438\\ .614 / \textbf{.419} / .395\\ .642 / .408 / .426\\ .600 / .406 / .460\\ .530 / .348 / .511\\ .554 / .305 / \textbf{.548}\\ .563 / .363 / .444\end{tabular}} & \multicolumn{1}{c|}{\begin{tabular}[c]{@{}c@{}}- / \textbf{.153} / .073\\ - / .119 / .006\\ - / .047 / -.025\\ - / .022 / -.010\\ - / -.028 / .043\\ - / -.005 / .061\\ - / -.009 / .073\\ - / .023 / .043\\ - / .044 / -.047\\ - / .008 / -.082\\ - / .023 / -.069\\ - / .102 / \textbf{.151}\end{tabular}} & \multicolumn{1}{c|}{\begin{tabular}[c]{@{}c@{}}.144 / .218 / .270\\ .155 / .208 / .319\\ .164 / .232 / .301\\ .176 / .241 / .326\\ .237 / .344 / .414\\ .309 / .397 / .471\\ .335 / .434 / .471\\ .389 / \textbf{.481} / .474\\ \textbf{.427} / .423 / \textbf{.479}\\ .354 / .333 / .433\\ .275 / .315 / .409\\ .271 / .398 / .264\end{tabular}} & \multicolumn{1}{c|}{\begin{tabular}[c]{@{}c@{}}.213 / .132 / .134\\ .255 / .129 / .234\\ .295 / .189 / .289\\ .307 / .254 / .286\\ \textbf{.305} / .321 / .351\\ .242 / .352 / .424\\ .237 / .404 / .441\\ .248 / .455 / .456\\ .250 / \textbf{.463} / .468\\ .275 / .414 / .497\\ .267 / .309 / \textbf{.500}\\ .270 / .389 / .257\end{tabular}} & \multicolumn{1}{c|}{\begin{tabular}[c]{@{}c@{}}.167 / .104 / .003\\
.255 / .164 / .076\\
.307 / .212 / .139\\
.394 / .276 / .184\\
.450 / .345 / .279\\
.468 / .361 / .277\\
\textbf{.479} / .364 / .280\\
.438 / \textbf{.430} / .297\\
.399 / .413 / .352\\
.282 / .331 / .407\\
.257 / .265 / \textbf{.444}\\
.335 / .341 / .386\end{tabular}} & \begin{tabular}[c]{@{}c@{}}.335 / .204 / .258\\ .374 / .198 / .245\\ .427 / .215 / .238\\ .492 / .257 / .287\\ .519 / .295 / .374\\ .516 / .338 / .438\\ .549 / .402 / \textbf{.439}\\ \textbf{.566} / \textbf{.427} / .430\\ .539 / .382 / .401\\ .515 / .362 / .369\\ .439 / .333 / .361\\ .518 / .368 / .290\end{tabular} & \begin{tabular}[c]{@{}c@{}}.281 / .204 / .308\\ .309 / .188 / .283\\ .370 / .218 / .292\\ .427 / .247 / .346\\ .465 / .275 / .453\\ .463 / .305 / \textbf{.503}\\ \textbf{.495} / .379 / .473\\ \textbf{.495} / \textbf{.400} / .466\\ .479 / .364 / .419\\ .461 / .313 / .405\\ .393 / .256 / .394\\ .482 / .345 / .287\end{tabular} & \multicolumn{1}{c|}{\begin{tabular}[c]{@{}c@{}}.261 / .214 / .253\\ .303 / .218 / .236\\ .360 / .242 / .241\\ .431 / .280 / .288\\ .456 / .318 / .373\\ .467 / .347 / \textbf{.432}\\ .523 / .429 / .430\\ .531 / \textbf{.451} / .427\\ \textbf{.534} / .405 / .404\\ .523 / .379 / .402\\ .461 / .330 / .401\\ .476 / .386 / .318\end{tabular}} & \begin{tabular}[c]{@{}c@{}}.160 / .143 / .145\\ .199 / .155 / .153\\ .290 / .170 / .171\\ .364 / .168 / .143\\ .396 / .192 / .165\\ .400 / .180 / .172\\ .429 / .262 / .191\\ .416 / \textbf{.291} / .197\\ .429 / .257 / .190\\ .418 / .226 / .191\\ .378 / .196 / \textbf{.215}\\ \textbf{.441} / .279 / .195\end{tabular} & \multicolumn{1}{c|}{\begin{tabular}[c]{@{}c@{}}.234 / .219 / .203\\ .288 / .213 / .235\\ .371 / .223 / .243\\ .463 / .322 / .264\\ .497 / .364 / .330\\ .532 / .374 / .367\\ \textbf{.547} / .437 / .375\\ .529 / \textbf{.499} / .373\\ .525 / .462 / .394\\ .531 / .425 / .411\\ .530 / .403 / \textbf{.432}\\ .466 / .488 / .379\end{tabular}} & \multicolumn{1}{c|}{\begin{tabular}[c]{@{}c@{}}.340 / -.100 / -.222\\ .540 / .263 / .338\\ .594 / .464 / .540\\ \textbf{.747} / .613 / .615\\ .720 / .662 / .600\\ .667 / .661 / .629\\ .645 / \textbf{.725} / .618\\ .654 / .715 / .638\\ .667 / .670 / \textbf{.646}\\ .625 / .656 / .613\\ .604 / .628 / .601\\ .656 / .689 / .500\end{tabular}} & \multicolumn{1}{c|}{\begin{tabular}[c]{@{}c@{}}.255 / .171 / .166 \\
.312 / .198 / .216 \\ 
.371 / .232 / .244 \\ 
.438 / .275 / .284 \\ 
.471 / .315 / .36 \\ 
.473 / .338 / \textbf{.398} \\ 
.494 / .390 / .393 \\ 
\textbf{.497} / \textbf{.421} / .396 \\ 
.486 / .391 / .388 \\ 
.450 / .346 / .387 \\ 
.405 / .303 / .392 \\ 
.449 / .371 / .316\end{tabular}} \\ \cline{2-15} 
\multicolumn{1}{c|}{} & \multicolumn{1}{c|}{PRT} & \multicolumn{1}{c|}{\begin{tabular}[c]{@{}c@{}}1\\ 2\\ 3\\ 4\\ 5\\ 6\\ 7\\ 8\\ 9\\ 10\\ 11\\ 12\end{tabular}} & \multicolumn{1}{c|}{\begin{tabular}[c]{@{}c@{}}.295 / .195 / .221\\ .409 / .271 / .382\\ .436 / .295 / .453\\ .467 / .290 / .487\\ .494 / .315 / .476\\ .516 / .353 / .447\\ .529 / \textbf{.383} / .462\\ .539 / \textbf{.383} / .464\\ \textbf{.549} / .358 / .437\\ .511 / .355 / .481\\ .452 / .342 / \textbf{.501}\\ .457 / .270 / .411\end{tabular}} & \multicolumn{1}{c|}{\begin{tabular}[c]{@{}c@{}}- / .289 / .303\\ - / .286 / .263\\ - / .277 / .271\\ - / .255 / .297\\ - / .232 / .322\\ - / .257 / .350\\ - / .304 / .349\\ - / .292 / .359\\ - / .311 / .319\\ - / .280 / .329\\ - / .298 / .308\\ - / \textbf{.380} / \textbf{.424}\end{tabular}} & \multicolumn{1}{c|}{\begin{tabular}[c]{@{}c@{}}.133 / .162 / .122\\ .217 / .198 / .125\\ .267 / .230 / .141\\ .297 / .285 / .204\\ .343 / .384 / .294\\ .379 / .421 / .357\\ .400 / .437 / .385\\ .398 / .468 / .402\\ .390 / \textbf{.469} / .477\\ .380 / .454 / .486\\ .412 / .430 / \textbf{.507}\\ \textbf{.422} / .436 / .369\end{tabular}} & \multicolumn{1}{c|}{\begin{tabular}[c]{@{}c@{}}.215 / .001 / .045\\ .274 / .006 / .066\\ \textbf{.301} / .012 / .078\\ .280 / .017 / .087\\ .233 / .060 / .129\\ .206 / .082 / .171\\ .178 / .008 / .184\\ .197 / .081 / .196\\ .201 / .096 / .247\\ .193 / \textbf{.133} / .223\\ .169 / .076 / \textbf{.245}\\ .158 / .193 / .020\end{tabular}} & \multicolumn{1}{c|}{\begin{tabular}[c]{@{}c@{}}.303 / .295 / .190\\
.407 / .397 / .328\\
.438 / .424 / .364\\
.455 / .446 / .388\\
.455 / .495 / .439\\
.451 / .524 / .449\\
.466 / .498 / .453\\
.453 / .514 / .463\\
\textbf{.476} / .501 / .503\\
.417 / .482 / .538\\
.422 / .489 / \textbf{.540}\\
.413 / \textbf{.543} / .505\end{tabular}} & \begin{tabular}[c]{@{}c@{}}.263 / .271 / .220\\ .304 / .279 / .216\\ .338 / .311 / .203\\ .398 / .329 / .246\\ .399 / .364 / .323\\ .391 / .359 / .365\\ \textbf{.411} / .379 / .358\\ .404 / \textbf{.393} / .375\\ .375 / .353 / .382\\ .349 / .376 / .409\\ .319 / .344 / \textbf{.412}\\ .400 / .391 / .321\end{tabular} & \begin{tabular}[c]{@{}c@{}}.206 / .149 / .305\\ .261 / .139 / .352\\ .305 / .191 / .405\\ .346 / .235 / .433\\ .395 / .327 / .509\\ .390 / .374 / .519\\ \textbf{.426} / \textbf{.447} / .510\\ .410 / .421 / \textbf{.531}\\ .402 / .404 / .471\\ .379 / .382 / .447\\ .317 / .335 / .439\\ .374 / .356 / .443\end{tabular} & \multicolumn{1}{c|}{\begin{tabular}[c]{@{}c@{}}.159 / .169 / .144\\ .196 / .161 / .153\\ .251 / .195 / .162\\ .306 / .250 / .234\\ .331 / .313 / .323\\ .331 / .365 / .384\\ \textbf{.380} / .413 / .384\\ \textbf{.380} / .411 / .396\\ .353 / .384 / .401\\ .335 / .366 / .431\\ .303 / .321 / \textbf{.438}\\ .347 / \textbf{.423} / .405\end{tabular}} & \begin{tabular}[c]{@{}c@{}}.032 / -.005 / .028\\ .122 / -.020 / .092\\ .250 / .042 / .111\\ .378 / .019 / .102\\ .440 / .096 / .137\\ .449 / .104 / .181\\ \textbf{.511} / .161 / .192\\ .449 / .227 / .292\\ .481 / \textbf{.243} / .351\\ .482 / .212 / .373\\ .448 / .197 / .360\\ .507 / .219 / \textbf{.387}\end{tabular} & \multicolumn{1}{c|}{\begin{tabular}[c]{@{}c@{}}.161 / .168 / .039\\ .349 / .215 / -.020\\ .365 / .294 / .005\\ .408 / .303 / .075\\ .466 / .367 / .189\\ .471 / .330 / .232\\ .501 / .371 / .236\\ .493 / .389 / \textbf{.246}\\ .485 / .380 / .239\\ .481 / .398 / .263\\ \textbf{.503} / .365 / .214\\ .444 / \textbf{.438} / .149\end{tabular}} & \multicolumn{1}{c|}{\begin{tabular}[c]{@{}c@{}}.383 / .017 / -.139\\ .582 / .192 / .140\\ .676 / .397 / .424\\ .691 / .525 / .544\\ .651 / .551 / .531\\ .637 / .556 / .475\\ .641 / .613 / .549\\ .664 / \textbf{.619} / .575\\ .671 / .606 / \textbf{.646}\\ .626 / .583 / .619\\ .602 / .550 / .620\\ \textbf{.712} / .524 / .558\end{tabular}} & \multicolumn{1}{c|}{\begin{tabular}[c]{@{}c@{}}.220 / .178 / .165 \\
.302 / .209 / .216\\ 
.348 / .253 / .253\\ 
.389 / .283 / .296\\ 
.408 / .337 / .357\\ 
.408 / .362 / .383\\ 
\textbf{.433} / .389 / .390\\ 
.426 / \textbf{.400} / .409\\ 
.422 / .385 / .418\\ 
.396 / .378 / .431\\ 
.371 / .350 / \textbf{.432}\\ 
.406 / .395 / .381\end{tabular}} \\ \cline{2-15} 
\multicolumn{1}{c|}{\multirow{2}{*}{\rotatebox{90}{\textbf{sense-based}} \vspace{-150pt}}} & \multicolumn{1}{c|}{AP} & \multicolumn{1}{c|}{\begin{tabular}[c]{@{}c@{}}1\\ 2\\ 3\\ 4\\ 5\\ 6\\ 7\\ 8\\ 9\\ 10\\ 11\\ 12\end{tabular}} & \multicolumn{1}{c|}{\begin{tabular}[c]{@{}c@{}}.129 / .220 / .032\\ .288 / .079 / -.128\\ .267 / .161 / .016\\ .353 / .330 / .087\\ \textbf{.432} / .221 / .322\\ .431 / .208 / .330\\ .144 / .362 / .321\\ .228 / \textbf{.418} / .175\\ .424 / .357 / .311\\ .233 / .317 / .289\\ .148 / .338 / \textbf{.374}\\ .289 / .181 / .278\end{tabular}} & \multicolumn{1}{c|}{\begin{tabular}[c]{@{}c@{}}- / -.011 / \textbf{.409}\\ - / .008 / .215\\ - / -.012 / .218\\ - / -.106 / .253\\ - / -.024 / .281\\ - / -.000 / .286\\ - / -.044 / .233\\ - / -.101 / .260\\ - / .120 / .153\\ - / .124 / .381\\ - / .132 / .266\\ - / \textbf{.277} / .398\end{tabular}} & \multicolumn{1}{c|}{\begin{tabular}[c]{@{}c@{}}-.108 / -.087 / -.040\\ .113 / -.131 / -.017\\ .007 / -.043 / .120\\ -.041 / .088 / .054\\ .143 / .235 / .196\\ .243 / .372 / .280\\ .284 / \textbf{.443} / .387\\ .417 / .353 / .393\\ .339 / .322 / .361\\ .393 / .328 / .334\\ .465 / .275 / \textbf{.435}\\ \textbf{.469} / .280 / .224\end{tabular}} & \multicolumn{1}{c|}{\begin{tabular}[c]{@{}c@{}}-.121 / -.021 / -.244\\ -.138 / -.141 / -.244\\ -.201 / -.117 / -.177\\ -.213 / -.131 / -.172\\ -.015 / -.083 / -.125\\ -.129 / -.040 / -.070\\ -.070 / -.031 / -.155\\ \textbf{.124} / .114 / -.082\\ .054 / .010 / -.195\\ -.023 / .061 / -.210\\ -.057 / \textbf{.175} / \textbf{.133}\\ -.090 / .023 / -.076\end{tabular}} & \multicolumn{1}{c|}{\begin{tabular}[c]{@{}c@{}}.168 / .233 / .172\\
.104 / .109 / .140\\
.161 / .142 / .063\\
.263 / .195 / \textbf{.266}\\
.247 / .319 / .162\\
.363 / .251 / .002\\
\textbf{.406} / .301 / .216\\
.384 / \textbf{.401} / .031\\
.270 / .296 / .157\\
.294 / .201 / .151\\
.351 / .310 / .039\\
.225 / .067 / .224\end{tabular}} & \begin{tabular}[c]{@{}c@{}}.050 / -.001 / -.154\\ -.127 / -.154 / -.036\\ -.006 / .007 / -.019\\ .093 / -.159 / -.042\\ .072 / -.085 / -.035\\ -.049 / -.111 / -.094\\ .082 / -.069 / \textbf{.067}\\ .058 / -.014 / -.073\\ .038 / .013 / -.081\\ \textbf{.126} / \textbf{.108} / .044\\ -.004 / .034 / -.069\\ .069 / .017 / -.068\end{tabular} & \begin{tabular}[c]{@{}c@{}}.132 / .108 / .060\\ .038 / .110 / .073\\ -.002 / .058 / .129\\ .045 / .096 / .104\\ .169 / .014 / .140\\ .173 / .093 / .176\\ .288 / \textbf{.235} / .084\\ .128 / .230 / .211\\ .072 / .149 / .232\\ .116 / .169 / .240\\ .068 / .141 / \textbf{.279}\\ \textbf{.279} / .086 / .209\end{tabular} & \multicolumn{1}{c|}{\begin{tabular}[c]{@{}c@{}}.098 / -.143 / .023\\ .096 / -.109 / -.025\\ .027 / -.130 / -.020\\ .168 / -.076 / .050\\ .081 / -.019 / .025\\.091 / .035 / .291\\ \textbf{.190} / \textbf{.158} / .131\\ .088 / .137 / .228\\ .098 / .055 / .011\\ .187 / .082 / .194\\ .157 / .113 / \textbf{.262}\\ .094 / -.116 / .130\end{tabular}} & \begin{tabular}[c]{@{}c@{}}-.104 / -.237 / -.019\\ .031 / -.230 / -.025\\ -.118 / .016 / -.060\\ -.281 / -.123 / -.016\\ -.318 / -.027 / .033\\ -.192 / -.076 / .031\\ -.257 / -.114 / -.051\\ -.165 / -.114 / -.109\\ -.016 / .005 / \textbf{.045}\\ .151 / -.127 / -.041\\ .021 / -.232 / -.211\\ \textbf{.314} / \textbf{.035} / -.100\end{tabular} & \multicolumn{1}{c|}{\begin{tabular}[c]{@{}c@{}}-.048 / .021 / -.239\\ -.039 / .104 / .028\\ -.051 / -.011 / .124\\ .257 / -.282 / .020\\ .323 / .143 / .149\\ \textbf{.440} / .206 / .131\\ .115 / .140 / -.130\\ -.029 / \textbf{.469} / \textbf{.256}\\ .092 / .198 / .031\\ .168 / .271 / .101\\ .090 / .146 / .062\\ .011 / -.090 / .030\end{tabular}} & \multicolumn{1}{c|}{\begin{tabular}[c]{@{}c@{}}.118 / -.179 / .110\\ .301 / -.058 / -.048\\ .189 / .221 / -.143\\ .360 / .322 / -.047\\ .251 / \textbf{.689} / .343\\ \textbf{.458} / .342 / .280\\ .292 / .226 / .344\\ .113 / .231 / .045\\ .423 / .404 / .245\\ .430 / .291 / .436\\ .322 / .223 / .243\\ .165 / .465 / \textbf{.448}\end{tabular}} & \multicolumn{1}{c|}{\begin{tabular}[c]{@{}c@{}}.060 / .011 / .012\\ 
.052 / -.030 / .006\\ 
.033 / .021 / .028\\ 
.113 / .014 / .064\\ 
.140 / .097 / .112\\
.166 / .099 / .132\\ 
.183 / .153 / .131\\ 
.148 / \textbf{.192} / .117\\ 
.157 / .158 / .104\\ 
\textbf{.197} / .158 / \textbf{.169}\\ 
.151 / .151 / .158\\ 
.179 / .077 / .142\end{tabular}} \\ \cline{2-15} 
\multicolumn{1}{c|}{} & \multicolumn{1}{c|}{WiDiD} & \multicolumn{1}{c|}{\begin{tabular}[c]{@{}c@{}}1\\ 2\\ 3\\ 4\\ 5\\ 6\\ 7\\ 8\\ 9\\ 10\\ 11\\ 12\end{tabular}} & \multicolumn{1}{c|}{\begin{tabular}[c]{@{}c@{}}.253 / .301 / .278\\ .434 / .261 / .065\\ .423 / .268 / .147\\ \textbf{.611} / .228 / .448\\ .527 / .078 / .393\\ .458 / .250 / .625\\ \textbf{.305} / .328 / .475\\ .449 / .312 / .411\\ .544 / \textbf{.509} / .567\\ .396 / .301 / .587\\ .299 / .218 / \textbf{.627}\\ .385 / .323 / .564\end{tabular}} & \multicolumn{1}{c|}{\begin{tabular}[c]{@{}c@{}}- / .028 / -.048\\ - / .018 / -.130\\ - / .026 / .019\\ - / .030 / .108\\ - / -.020 / -.037\\ - / -.030 / -.050\\ - / \textbf{.139} / .106\\ - / .091 / .038\\ - / -.066 / .104\\ - / -.024 / \textbf{.187}\\ - / -.064 / -.111\\ - / -.039 / -.064\end{tabular}} & \multicolumn{1}{c|}{\begin{tabular}[c]{@{}c@{}}.147 / .204 / .219\\ .106 / .143 / .292\\ .115 / .120 / .474\\ .126 / .067 / .424\\ .190 / .173 / \textbf{.509}\\ .293 / .294 / .433\\ .235 / .253 / .514\\ .344 / .341 / .565\\ .353 / .299 / .573\\ .315 / \textbf{.407} / .477\\ .258 / .381 / .486\\ \textbf{.355} / .312 / .499\end{tabular}} & \multicolumn{1}{c|}{\begin{tabular}[c]{@{}c@{}}.120 / .052 / -.062\\ -.041 / .015 / -.118\\ .198 / .029 / .106\\ .176 / -.130 / .312\\ .151 / -.074 / .300\\ .211 / .148 / .335\\ \textbf{.295} / .198 / \textbf{.414}\\ .071 / \textbf{.354} / .321\\ .184 / .319 / .203\\ .145 / .233 / .148\\ .172 / .128 / .343\\ .106 / .195 / .129\end{tabular}} & \multicolumn{1}{c|}{\begin{tabular}[c]{@{}c@{}}.132 / .051 / -.015\\
.103 / .105 / .110\\
.228 / .108 / .118\\
.292 / .175 / .221\\
.356 / .295 / .310\\
.382 / .387 / .346\\
.382 / .318 / .324\\
.340 / .371 / .395\\
.324 / \textbf{.450} / .372\\
.306 / .388 / \textbf{.471}\\
\textbf{.424} / .432 / .464\\
.383 / .343 / .459\end{tabular}} & \begin{tabular}[c]{@{}c@{}}.159 / .047 / .125\\ .209 / -.046 / .274\\ \textbf{.251} / -.073 / \textbf{.345}\\ .091 / -.039 / .332\\ -.034 / .023 / .259\\ .094 / .063 / .184\\ .017 / .032 / .292\\ .000 / -.008 / .105\\ -.002 / .075 / .108\\ .011 / .087 / .270\\ .134 / \textbf{.152} / .220\\ .135 / -.068 / .268\end{tabular} & \begin{tabular}[c]{@{}c@{}}.108 / .073 / .197\\ .076 / .180 / .060\\ .091 / .113 / .184\\ .010 / .041 / .307\\ .071 / .076 / .314\\ .141 / .066 / .210\\ .203 / \textbf{.285} / .152\\ .284 / .260 / .243\\ .083 / .076 / .171\\ \textbf{.302} / .090 / .308\\ .234 / .120 / \textbf{.334}\\ .102 / .160 / .216\end{tabular} & \multicolumn{1}{c|}{\begin{tabular}[c]{@{}c@{}}.090 / -.036 / .051\\ .212 / -.038 / -.008\\ .233 / .077 / .153\\ .157 / -.053 / .059\\ .205 / .137 / .202\\ .182 / \textbf{.288} / .264\\ .216 / .188 / \textbf{.458}\\ .025 / .203 / .267\\ .205 / .205 / .388\\ .060 / .172 / .328\\ .185 / .087 / .312\\ \textbf{.243} / .142 / .342\end{tabular}} & \begin{tabular}[c]{@{}c@{}}.356 / .150 / .090\\ .285 / -.030 / .085\\ .229 / -.102 / .074\\ .242 / .038 / .002\\ \textbf{.297} / .100 / .023\\ .261 / -.080 / .215\\ .244 / .119 / .247\\ .221 / .226 / .262\\ .183 / .063 / .174\\ .155 / .179 / .234\\ .218 / .195 / \textbf{.345}\\ .233 / \textbf{.241} / .226\end{tabular} & \multicolumn{1}{c|}{\begin{tabular}[c]{@{}c@{}}.120 / .127 / .154\\ .161 / .103 / .214\\ .239 / .064 / .204\\ .340 / .152 / .062\\ .380 / .156 / .316\\ .428 / .295 / .102\\ .397 / .195 / -.034\\ .449 / \textbf{.428} / .155\\ .390 / .118 / .149\\ \textbf{.488} / .175 / .275\\ .296 / .291 / \textbf{.438}\\ .087 / .290 / .349\end{tabular}} & \multicolumn{1}{c|}{\begin{tabular}[c]{@{}c@{}}.122 / .026 / .160\\ .371 / -.013 / .063\\ .256 / .114 / .349\\ .388 / .279 / \textbf{.417}\\ .524 / .193 / .217\\ .446 / .271 / .335\\ .338 / .298 / .293\\ .475 / .325 / .286\\ .404 / .347 / .328\\ .428 / \textbf{.355} / .383\\ \textbf{.539} / .277 / .372\\ .533 / .338 / .382\end{tabular}} & \multicolumn{1}{c|}{\begin{tabular}[c]{@{}c@{}}.146 / .074 / .103\\ 
.175 / .060 / .094\\
.216 / .065 / .203\\ 
.200 / .054 / .244\\
.218 / .112 / .265\\ 
.252 / .185 / .269\\
.237 / .211 / .304\\ 
.224 / \textbf{.242} / .271\\ 
.222 / .212 / .280\\
.224 / .204 / .339\\ 
\textbf{.260} / .199 / \textbf{.345}\\
.239 / .181 / .314 \end{tabular}} \\ \cline{2-15} 
\end{tabular}%
}
\caption{\textbf{Comprehensive evaluation of standard approaches to GCD} by using the layers 1-12 of \textbf{BERT / mBERT / XLM-R}. Top score for each approach, model, and benchmark in \textbf{bold}. Avg is the weighted average score based on the number of targets in each benchmark. }
\label{tab:all-layers}
\end{table}
\end{landscape}

\begin{minipage}{\textwidth}
\centering
\resizebox{\textwidth}{!}{%
\begin{tabular}{ccccccccccccc}
\multicolumn{1}{l}{} &  & \textbf{EN} & \textbf{LA} & \textbf{DE} & \textbf{SV} & \textbf{ES} & \multicolumn{3}{c}{\textbf{RU}} & \multicolumn{2}{c}{\textbf{NO}} & \textbf{ZH} \\ \cline{3-13} 
\multicolumn{1}{l}{} & \multicolumn{1}{c|}{} & \multicolumn{1}{c|}{\textit{$C_1 - C_2$}} & \multicolumn{1}{c|}{\textit{$C_1 - C_2$}} & \multicolumn{1}{c|}{\textit{$C_1 - C_2$}} & \multicolumn{1}{c|}{\textit{$C_1 - C_2$}} & \multicolumn{1}{c|}{\textit{$C_1 - C_2$}} & \textit{$C_1 - C_2$} & \textit{$C_2 - C_3$} & \multicolumn{1}{c|}{\textit{$C_1 - C_3$}} & \textit{$C_1 - C_2$} & \multicolumn{1}{c|}{\textit{$C_2 - C_3$}} & \multicolumn{1}{c|}{\textit{$C_1 - C_2$}} \\ \cline{2-13} 
\multicolumn{1}{c|}{\multirow{2}{*}{\rotatebox{90}{\textbf{form-based}}}} & \multicolumn{1}{c|}{\rotatebox{90}{\textbf{APD}}} & \multicolumn{1}{c|}{\begin{tabular}[c]{@{}c@{}}XL-L. : .757\\ \citeauthor{cassotti2023xl}\\ \cline{1-1} BERT: .706\\ \citeauthor{zhou2023finer}\end{tabular}} & \multicolumn{1}{c|}{\begin{tabular}[c]{@{}c@{}}XL-L. : -.056\\ \citeauthor{cassotti2023xl}\\ \cline{1-1} mBERT: .443\\ \citeauthor{pomsl2020circe}\end{tabular}} & \multicolumn{1}{c|}{\begin{tabular}[c]{@{}c@{}}XL-L. : .877\\ \citeauthor{cassotti2023xl}\\ \cline{1-1} BERT: .731\\ \citeauthor{laicher2021explaining}\end{tabular}} & \multicolumn{1}{c|}{\begin{tabular}[c]{@{}c@{}}XL-L. .754\\ \citeauthor{cassotti2023xl}\\ \cline{1-1} BERT: .602\\ \citeauthor{laicher2021explaining}\end{tabular}} & \multicolumn{1}{c|}{\begin{tabular}[c]{@{}c@{}}n.a.\\ \cline{1-1} n.a.\end{tabular}} & \begin{tabular}[c]{@{}c@{}}XL-L. : .799\\ \citeauthor{cassotti2023xl}\\ \cline{1-1} XLM-R: .372\\ \citeauthor{giulianelli2022fire}\end{tabular} & \begin{tabular}[c]{@{}c@{}}XL-L. : .833\\ \citeauthor{cassotti2023xl}\\ \cline{1-1}XLM-R: .480 \\ \citeauthor{giulianelli2022fire}\end{tabular} & \multicolumn{1}{c|}{\begin{tabular}[c]{@{}c@{}}XL-L. : .842\\ \citeauthor{cassotti2023xl}\\ \cline{1-1} XLM-R: .457\\ \citeauthor{giulianelli2022fire}\end{tabular}} & \begin{tabular}[c]{@{}c@{}}XL-L. : .757\\ \citeauthor{cassotti2023xl}\\ \cline{1-1} XLM-R: .389\\ \citeauthor{giulianelli2022fire}\end{tabular} & \multicolumn{1}{c|}{\begin{tabular}[c]{@{}c@{}}XL-L. : .757\\ \citeauthor{cassotti2023xl}\\ \cline{1-1}XLM-R: .387\\ \citeauthor{giulianelli2022fire}\end{tabular}} & \multicolumn{1}{c|}{\begin{tabular}[c]{@{}c@{}}n.a.\\ \cline{1-1} n.a.\end{tabular}} \\ \cline{2-13} 
\multicolumn{1}{c|}{} & \multicolumn{1}{c|}{\rotatebox{90}{PRT}} & \multicolumn{1}{c|}{\begin{tabular}[c]{@{}c@{}}BERT: .531 \\ \citeauthor{zhou2023finer}\\ \cline{1-1} BERT: .467\\ \citeauthor{rosin2022time}\end{tabular}} & \multicolumn{1}{c|}{\begin{tabular}[c]{@{}c@{}}n.a.\\ \cline{1-1} mBERT: .561\\ \citeauthor{kutuzov2020uio}\end{tabular}} & \multicolumn{1}{c|}{\begin{tabular}[c]{@{}c@{}}n.a.\\ \cline{1-1} BERT: .755\\ \citeauthor{laicher2021explaining}\end{tabular}} & \multicolumn{1}{c|}{\begin{tabular}[c]{@{}c@{}}n.a.\\ \cline{1-1} BERT: .392\\ \citeauthor{zhou2020temporalteller}\end{tabular}} & \multicolumn{1}{c|}{\begin{tabular}[c]{@{}c@{}}n.a.\\ \cline{1-1} n.a.\end{tabular}} & \begin{tabular}[c]{@{}c@{}}n.a.\\ \cline{1-1} XLM-R: .294\\ \citeauthor{giulianelli2022fire}\end{tabular} & \begin{tabular}[c]{@{}c@{}}n.a.\\ \cline{1-1} XLM-R: .313\\ \citeauthor{giulianelli2022fire}\end{tabular} & \multicolumn{1}{c|}{\begin{tabular}[c]{@{}c@{}}n.a.\\ \cline{1-1} XLM-R: .313\\ \citeauthor{giulianelli2022fire}\end{tabular}} & \begin{tabular}[c]{@{}c@{}}n.a.\\ \cline{1-1} XLM-R: .378\\ \citeauthor{giulianelli2022fire}\end{tabular} & \multicolumn{1}{c|}{\begin{tabular}[c]{@{}c@{}}n.a.\\ \cline{1-1} XLM-R: .270\\ \citeauthor{giulianelli2022fire}\end{tabular}} & \multicolumn{1}{c|}{\begin{tabular}[c]{@{}c@{}}n.a.\\ \cline{1-1} n.a.\end{tabular}} \\ \cline{2-13} 
\multicolumn{1}{c|}{\multirow{2}{*}{\rotatebox{90}{\textbf{sense-based}}}} & \multicolumn{1}{c|}{\rotatebox{90}{AP+JSD}} & \multicolumn{1}{c|}{\begin{tabular}[c]{@{}c@{}}n.a. \\ \cline{1-1}BERT: .436 \\ \citeauthor{martinc2020discovery}\end{tabular}} & \multicolumn{1}{c|}{\begin{tabular}[c]{@{}c@{}}n.a. \\ \cline{1-1}mBERT: .481 \\ \citeauthor{martinc2020discovery}\end{tabular}} & \multicolumn{1}{c|}{\begin{tabular}[c]{@{}c@{}}n.a. \\ \cline{1-1}BERT: .583 \\ \citeauthor{montariol2021scalable}\end{tabular}} & \multicolumn{1}{c|}{\begin{tabular}[c]{@{}c@{}}n.a. \\ \cline{1-1}BERT: .343 \\ \citeauthor{martinc2020discovery}\end{tabular}} & \multicolumn{1}{c|}{\begin{tabular}[c]{@{}c@{}}n.a.\\ \cline{1-1} n.a.\end{tabular}} & \begin{tabular}[c]{@{}c@{}}n.a.\\ \cline{1-1} n.a.\end{tabular} & \begin{tabular}[c]{@{}c@{}}n.a.\\ \cline{1-1} n.a.\end{tabular} & \multicolumn{1}{c|}{\begin{tabular}[c]{@{}c@{}}n.a.\\ \cline{1-1} n.a.\end{tabular}} & \begin{tabular}[c]{@{}c@{}}n.a.\\ \cline{1-1} n.a.\end{tabular} & \multicolumn{1}{c|}{\begin{tabular}[c]{@{}c@{}}n.a.\\ \cline{1-1} n.a.\end{tabular}} & \multicolumn{1}{c|}{\begin{tabular}[c]{@{}c@{}}n.a.\\ \cline{1-1} n.a.\end{tabular}} \\ \cline{2-13} 
\multicolumn{1}{c|}{} & \multicolumn{1}{c|}{\rotatebox{90}{\textbf{WiDiD}}} & \multicolumn{1}{c|}{\begin{tabular}[c]{@{}c@{}}n.a. \\ \cline{1-1}BERT: .651 \\ \citeauthor{periti2023word}\end{tabular}} & \multicolumn{1}{c|}{\begin{tabular}[c]{@{}c@{}}n.a. \\ \cline{1-1}XLM-R: - .096 \\ \citeauthor{periti2023word}\end{tabular}} & \multicolumn{1}{c|}{\begin{tabular}[c]{@{}c@{}}n.a. \\ \cline{1-1}XLM-R: .527 \\ \citeauthor{periti2023word}\end{tabular}} & \multicolumn{1}{c|}{\begin{tabular}[c]{@{}c@{}}n.a. \\ \cline{1-1}XLM-R: .499 \\ \citeauthor{periti2023word}\end{tabular}} & \multicolumn{1}{c|}{\begin{tabular}[c]{@{}c@{}}n.a. \\ \cline{1-1}BERT: .544 \\ \citeauthor{periti2023word}\end{tabular}} & \begin{tabular}[c]{@{}c@{}}n.a. \\ \cline{1-1}mBERT: .273 \\ \citeauthor{periti2023word}\end{tabular} & \begin{tabular}[c]{@{}c@{}}n.a. \\ \cline{1-1}mBERT: .393 \\ \citeauthor{periti2023word}\end{tabular} & \multicolumn{1}{c|}{\begin{tabular}[c]{@{}c@{}}n.a. \\ \cline{1-1}mBERT: .407\\ \citeauthor{periti2023word}\end{tabular}} & \begin{tabular}[c]{@{}c@{}}n.a.\\ \cline{1-1} n.a.\end{tabular} & \multicolumn{1}{c|}{\begin{tabular}[c]{@{}c@{}}n.a.\\ \cline{1-1} n.a.\end{tabular}} & \multicolumn{1}{c|}{\begin{tabular}[c]{@{}c@{}}n.a.\\ \cline{1-1} n.a.\end{tabular}} \\ \cline{2-13} 
\end{tabular}%
}
\captionof{table}{\textbf{State-of-the-art performance for GCD}: Top Spearman correlations obtained across benchmarks by form- and sense-based approaches. For each approach, we report correlation for both \textit{supervised} (above the line) and \textit{unsupervised} (below the line) settings.}
\label{tab:sota}
\end{minipage}

\vspace{1cm}

\begin{minipage}{\textwidth}
\centering
\resizebox{\textwidth}{!}{%
\begin{tabular}{cccccccccccc}
 & \textbf{EN} & \textbf{LA} & \textbf{DE} & \textbf{SV} & \textbf{ES} & \multicolumn{3}{c}{\textbf{RU}} & \multicolumn{2}{c}{\textbf{NO}} & \textbf{ZH} \\ \cline{2-12} 
\multicolumn{1}{c|}{} & \multicolumn{1}{c|}{\textit{$C_1 - C_2$}} & \multicolumn{1}{c|}{\textit{$C_1 - C_2$}} & \multicolumn{1}{c|}{\textit{$C_1 - C_2$}} & \multicolumn{1}{c|}{\textit{$C_1 - C_2$}} & \multicolumn{1}{c|}{\textit{$C_1 - C_2$}} & \textit{$C_1 - C_2$} & \textit{$C_2 - C_3$} & \multicolumn{1}{c|}{\textit{$C_1 - C_3$}} & \textit{$C_1 - C_2$} & \multicolumn{1}{c|}{\textit{$C_2 - C_3$}} & \multicolumn{1}{c|}{\textit{$C_1 - C_2$}} \\ \hline
\multicolumn{1}{|c|}{\begin{tabular}[c]{@{}c@{}}Time \\ periods\end{tabular}} & \multicolumn{1}{c|}{\begin{tabular}[c]{@{}c@{}}$C_1$: 1810 -- 1860\\ $C_2$: 1960 -- 2010\end{tabular}} & \multicolumn{1}{c|}{\begin{tabular}[c]{@{}c@{}}$C_1$: 200 -- 0\\ $C_2$: 0 -- 2000\end{tabular}} & \multicolumn{1}{c|}{\begin{tabular}[c]{@{}c@{}}$C_1$: 1800 -- 1899\\ $C_2$: 1946 -- 1990\end{tabular}} & \multicolumn{1}{c|}{\begin{tabular}[c]{@{}c@{}}$C_1$: 1790 -- 1830\\ $C_2$: 1895 -- 1903\end{tabular}} & \multicolumn{1}{c|}{\begin{tabular}[c]{@{}c@{}}$C_1$: 1810 -- 1906\\ $C_2$: 1994 -- 2020\end{tabular}} & \begin{tabular}[c]{@{}c@{}}$C_1$: 1700 -- 1916\\ $C_2$: 1918 -- 1990\end{tabular} & \begin{tabular}[c]{@{}c@{}}$C_2$: 1918 -- 1990\\ $C_3$: 1992 --2016\end{tabular} & \multicolumn{1}{c|}{\begin{tabular}[c]{@{}c@{}}$C_1$: 1700 -- 1916\\ $C_3$: 1992 --2016\end{tabular}} & \begin{tabular}[c]{@{}c@{}}$C_1$: 1929 --1965\\ $C_2$: 1970 -- 2013\end{tabular} & \multicolumn{1}{c|}{\begin{tabular}[c]{@{}c@{}}$C_1$: 1980 -- 1990\\ $C_2$: 2012 -- 2019\end{tabular}} & \multicolumn{1}{c|}{\begin{tabular}[c]{@{}c@{}}$C_1$: 1954 -- 1978\\ $C_2$: 1979 -- 2003\end{tabular}} \\ \hline
\multicolumn{1}{|c|}{\begin{tabular}[c]{@{}c@{}}Diachronic\\ Corpus\end{tabular}} & \multicolumn{1}{c|}{\begin{tabular}[c]{@{}c@{}}$C_1$: CCOHA\\ $C_2$: CCOHA\end{tabular}} & \multicolumn{1}{c|}{\begin{tabular}[c]{@{}c@{}}$C_1$: LatinISE\\ $C_2$: LatinISE\end{tabular}} & \multicolumn{1}{c|}{\begin{tabular}[c]{@{}c@{}}$C_1$: DTA \\ $C_2$: BZ+ND\end{tabular}} & \multicolumn{1}{c|}{\begin{tabular}[c]{@{}c@{}}$C_1$: Kubhist\\ $C_2$: Kubhist\end{tabular}} & \multicolumn{1}{c|}{\begin{tabular}[c]{@{}c@{}}$C_1$: PG\\ $C_2$: TED2013, \\ NC\\ MultiUN\\ Europarl\end{tabular}} & \begin{tabular}[c]{@{}c@{}}$C_1$: RNC\\ $C_2$: RNC\\ $C_3$: RNC\end{tabular} & \begin{tabular}[c]{@{}c@{}}$C_1$: RNC\\ $C_2$: RNC\\ $C_3$: RNC\end{tabular} & \multicolumn{1}{c|}{\begin{tabular}[c]{@{}c@{}}$C_1$: RNC\\ $C_2$: RNC\\ $C_3$: RNC\end{tabular}} & \begin{tabular}[c]{@{}c@{}}$C_1$: NBdigital\\ $C_2$: NBdigital\end{tabular} & \multicolumn{1}{c|}{\begin{tabular}[c]{@{}c@{}}$C_1$: NBdigital\\ $C_2$: NAK\end{tabular}} & \multicolumn{1}{c|}{\begin{tabular}[c]{@{}c@{}}$C_1$: People's Daily\\ $C_2$: People's Daily\end{tabular}} \\ \hline
\multicolumn{1}{|c|}{\# targets} & \multicolumn{1}{c|}{46} & \multicolumn{1}{c|}{40} & \multicolumn{1}{c|}{50} & \multicolumn{1}{c|}{44} & \multicolumn{1}{c|}{100} & 111 & 111 & \multicolumn{1}{c|}{111} & 40 & \multicolumn{1}{c|}{40} & \multicolumn{1}{c|}{40} \\ \hline
\multicolumn{1}{|c|}{\begin{tabular}[c]{@{}c@{}}Benchmark\\ version\end{tabular}} & \multicolumn{1}{c|}{\begin{tabular}[c]{@{}c@{}}version 2.0.1\\\citeauthor{schlechtweg2022dwugen}\end{tabular}} & \multicolumn{1}{c|}{\begin{tabular}[c]{@{}c@{}}version 1\\\citeauthor{barbara2021dwugla}\end{tabular}} & \multicolumn{1}{c|}{\begin{tabular}[c]{@{}c@{}}version 2.3.0\\\citeauthor{schlechtweg2022dwugde}\end{tabular}} & \multicolumn{1}{c|}{\begin{tabular}[c]{@{}c@{}}version 2.0.1\\\citeauthor{tahmasebi2022dwugsv}\end{tabular}} & \multicolumn{1}{c|}{\begin{tabular}[c]{@{}c@{}}version 4.0.0\\\citeauthor{zamora2022dwuges}\end{tabular}} & \multicolumn{3}{c|}{\begin{tabular}[c]{@{}c@{}}version 1\\\citeauthor{kutuzov2022dwugru}\end{tabular}} & \multicolumn{2}{c|}{\begin{tabular}[c]{@{}c@{}}version 1\\\citeauthor{kutuzov2022dwugno}\end{tabular}} & \multicolumn{1}{c|}{\begin{tabular}[c]{@{}c@{}}version 1\\\citeauthor{chen2023chiwug}\end{tabular}} \\ \hline
\end{tabular}%
}
\captionof{table}{\textbf{LSC benchmark for Graded Change Detection}. Overview of time periods, diachronic corpus composition, number of targets, and benchmark versions used in this study.}
\label{tab:benchmarks}
\end{minipage}

\vspace{1cm}
\begin{minipage}{\textwidth}
\centering
\resizebox{\textwidth}{!}{
\begin{tabular}{ccccc}
\textbf{} & \textbf{BERT} & \textbf{mBERT} & \textbf{XLM-R} & \textbf{XL-LEXEME} \\ \cline{2-5} 
\multicolumn{1}{c|}{\textbf{English}} & \texttt{bert-base-uncased} & \texttt{bert-base-multilingual-cased} & \texttt{xlm-roberta-base} & \multicolumn{1}{c|}{\texttt{pierluigic/xl-lexeme}} \\ \cline{2-5} 
\multicolumn{1}{c|}{\textbf{Latin}} & - & \texttt{bert-base-multilingual-cased} & \texttt{xlm-roberta-base} & \multicolumn{1}{c|}{\texttt{pierluigic/xl-lexeme}} \\ \cline{2-5} 
\multicolumn{1}{c|}{\textbf{German}} & \texttt{bert-base-german-cased} & \texttt{bert-base-multilingual-cased} & \texttt{xlm-roberta-base} & \multicolumn{1}{c|}{\texttt{pierluigic/xl-lexeme}} \\ \cline{2-5} 
\multicolumn{1}{c|}{\textbf{Swedish}} & \texttt{af-ai-center/bert-base-swedish-uncased} & \texttt{bert-base-multilingual-cased} & \texttt{xlm-roberta-base} & \multicolumn{1}{c|}{\texttt{pierluigic/xl-lexeme}} \\ \cline{2-5} 
\multicolumn{1}{c|}{\textbf{Spanish}} & \texttt{dccuchile/bert-base-spanish-wwm-uncased} & \texttt{bert-base-multilingual-cased} & \texttt{xlm-roberta-base} & \multicolumn{1}{c|}{\texttt{pierluigic/xl-lexeme}} \\ \cline{2-5} 
\multicolumn{1}{c|}{\textbf{Russian}} & \texttt{DeepPavlov/rubert-base-cased} & \texttt{bert-base-multilingual-cased} & \texttt{xlm-roberta-base} & \multicolumn{1}{c|}{\texttt{pierluigic/xl-lexeme}} \\ \cline{2-5} 
\multicolumn{1}{c|}{\textbf{Norwegian}} & \texttt{NbAiLab/nb-bert-base} & \texttt{bert-base-multilingual-cased} & \texttt{xlm-roberta-base} & \multicolumn{1}{c|}{\texttt{pierluigic/xl-lexeme}} \\ \cline{2-5} 
\multicolumn{1}{c|}{\textbf{Chinese}} & \texttt{bert-base-chinese} & \texttt{bert-base-multilingual-cased} & \texttt{xlm-roberta-base} & \multicolumn{1}{c|}{\texttt{pierluigic/xl-lexeme}} \\ \cline{2-5} 
\end{tabular}%
}
\captionof{table}{BERT, mBERT, XLM-R, and XL-LEXEME models employed in our evaluation. All models are available at \url{huggingface.co}.}
\label{tab:models-evaluated}
\end{minipage}

\newpage

\newpage

\newpage

\begin{landscape}
\begin{table}]
\centering
\resizebox{1.5\textwidth}{!}{%
\begin{tabular}{cccccccccccccc}
 &  & & \textbf{EN} & \textbf{LA} & \textbf{DE} & \textbf{SV} & \textbf{ES} & \multicolumn{3}{c}{\textbf{RU}} & \multicolumn{2}{c}{\textbf{NO}} & \textbf{ZH} \\ \cline{4-14} 
 &  & \multicolumn{1}{c|}{} & \multicolumn{1}{c|}{\textit{$C_1 - C_2$}} & \multicolumn{1}{c|}{\textit{$C_1 - C_2$}} & \multicolumn{1}{c|}{\textit{$C_1 - C_2$}} & \multicolumn{1}{c|}{\textit{$C_1 - C_2$}} & \multicolumn{1}{c|}{\textit{$C_1 - C_2$}} & \textit{$C_1 - C_2$} & \textit{$C_2 - C_3$} & \multicolumn{1}{c|}{\textit{$C_1 - C_3$}} & \textit{$C_1 - C_2$} & \multicolumn{1}{c|}{\textit{$C_2 - C_3$}} & \multicolumn{1}{c|}{\textit{$C_1 - C_2$}} \\ \cline{2-14} 
\multicolumn{1}{c|}{\multirow{2}{*}{\rotatebox{90}{\textbf{form-based} }}} & \multicolumn{1}{c|}{APD} & \multicolumn{1}{c|}{\begin{tabular}[c]{@{}c@{}}BERT\\ mBERT\\ XLM-R\end{tabular}} & \multicolumn{1}{c|}{\begin{tabular}[c]{@{}c@{}}\textbf{.692} / .566 \textit{(.563)}\\ .466 / .365 \textit{(.363)}\\ .579 / .518 \textit{(.444)}\end{tabular}} & \multicolumn{1}{c|}{\begin{tabular}[c]{@{}c@{}}/\\ .136 / .034 \textit{(.102)}\\ .080 / -.072 \textit{\textbf{(.151)}}\end{tabular}} & \multicolumn{1}{c|}{\begin{tabular}[c]{@{}c@{}}.412 / .349 \textit{(.271)}\\ .468 / .370 \textit{(.398)}\\ \textbf{.496} / .438 \textit{(.264)}\end{tabular}} & \multicolumn{1}{c|}{\begin{tabular}[c]{@{}c@{}}.325 / .272 \textit{(.270)}\\ .486 / .398 \textit{(.389)}\\ \textbf{.496} / \textbf{.496} \textit{(.257)}\end{tabular}} & \multicolumn{1}{c|}{\begin{tabular}[c]{@{}c@{}}\textbf{.488} / .310 \textit{(.335)}\\ .423 / .351 \textit{(.341)}\\ .443 / .398 \textit{(.386)}\end{tabular}} & \begin{tabular}[c]{@{}c@{}}\textbf{.573} / .537 \textit{(.518)}\\ .419 / .365 \textit{(.368)}\\ .441 / .368 \textit{(.290)}\end{tabular} & \begin{tabular}[c]{@{}c@{}}\textbf{.506} / .477 \textit{(.482)}\\ .393 / .324 \textit{(.345)}\\ .491 / .404 \textit{(.287)}\end{tabular} & \multicolumn{1}{c|}{\begin{tabular}[c]{@{}c@{}}\textbf{.546} / .522 \textit{(.476)}\\ .443 / .386 \textit{(.386)}\\ .432 / .397 \textit{(.318)}\end{tabular}} & \begin{tabular}[c]{@{}c@{}}\textbf{.463} / .457 \textit{(.441)}\\ .320 / .248 \textit{(.279)}\\ .215 / .180 \textit{(.195)}\end{tabular} & \multicolumn{1}{c|}{\begin{tabular}[c]{@{}c@{}}\textbf{.556} / .521 \textit{(.466)}\\ .496 / .429 \textit{(.488)}\\ .421 / .418 \textit{(.379)}\end{tabular}} & \multicolumn{1}{c|}{\begin{tabular}[c]{@{}c@{}}\textbf{.760} / .658 \textit{(.656)}\\ .739 / .674 \textit{(.689)}\\ .675 / .627 \textit{(.500)}\end{tabular}} \\ \cline{2-14} 
\multicolumn{1}{c|}{} & \multicolumn{1}{c|}{PRT} & \multicolumn{1}{c|}{\begin{tabular}[c]{@{}c@{}}BERT\\ mBERT\\ XLM-R\end{tabular}} & \multicolumn{1}{c|}{\begin{tabular}[c]{@{}c@{}}\textbf{.550} / .520 \textit{(.457)}\\ .382 / .339 \textit{(.270)}\\ .513 / .476 \textit{(.411)}\end{tabular}} & \multicolumn{1}{c|}{\begin{tabular}[c]{@{}c@{}}/\\ .352 / .305 \textit{(.380)}\\ .365 / .312 \textit{\textbf{(.424)}}\end{tabular}} & \multicolumn{1}{c|}{\begin{tabular}[c]{@{}c@{}}.421 / .397 \textit{(.422)}\\ .467 / .454 \textit{(.436)}\\ \textbf{.497} / .486 \textit{(.369)}\end{tabular}} & \multicolumn{1}{c|}{\begin{tabular}[c]{@{}c@{}}\textbf{.293} / .170 \textit{(.158)}\\ .132 / .105 \textit{(.193)}\\ .253 / .236 \textit{(.020)}\end{tabular}} & \multicolumn{1}{c|}{\begin{tabular}[c]{@{}c@{}}.478 / .441 \textit{(.413)}\\ \textbf{.555} / .514 \textit{(.543)}\\ .538 / .522 \textit{(.505)}\end{tabular}} & \begin{tabular}[c]{@{}c@{}}\textbf{.425} / .368 \textit{(.400)}\\ .411 / .373 \textit{(.391)}\\ .409 / .402 \textit{(.320)}\end{tabular} & \begin{tabular}[c]{@{}c@{}}.418 / .374 \textit{(.374)}\\ .442 / .386 \textit{(.356)}\\ \textbf{.530} / .453 \textit{(.443)}\end{tabular} & \multicolumn{1}{c|}{\begin{tabular}[c]{@{}c@{}}.383 / .346 \textit{(.347)}\\ .434 / .367 \textit{(.423)}\\ \textbf{.449} / .435 \textit{(.405)}\end{tabular}} & \begin{tabular}[c]{@{}c@{}}\textbf{.538} / .513 \textit{(.507)}\\ .256 / .228 \textit{(.219)}\\ .384 / .384 \textit{(.387)}\end{tabular} & \multicolumn{1}{c|}{\begin{tabular}[c]{@{}c@{}}\textbf{.513} / .481 \textit{(.444)}\\ .432 / .405 \textit{(.438)}\\ .270 / .220 \textit{(.149)}\end{tabular}} & \multicolumn{1}{c|}{\begin{tabular}[c]{@{}c@{}}.706 / .649 \textit{(\textbf{.712})}\\ .648 / .588 \textit{(.524)}\\ .642 / .627 \textit{(.558)}\end{tabular}} \\ \cline{2-14} 
\multicolumn{1}{c|}{\multirow{2}{*}{\rotatebox{90}{\textbf{sense-based}}}} & \multicolumn{1}{c|}{AP} & \multicolumn{1}{c|}{\begin{tabular}[c]{@{}c@{}}BERT\\ mBERT\\ XLM-R\end{tabular}} & \multicolumn{1}{c|}{\begin{tabular}[c]{@{}c@{}}.464 / .245 \textit{(.289)}\\ \textbf{.501} / .313 \textit{(.181)}\\ .473 / .340 \textit{(.278)}\end{tabular}} & \multicolumn{1}{c|}{\begin{tabular}[c]{@{}c@{}}/\\ .326 / .179 \textit{(.277)}\\ \textbf{.482} / .398 \textit{(.398)}\end{tabular}} & \multicolumn{1}{c|}{\begin{tabular}[c]{@{}c@{}}\textbf{.520} / .435 \textit{(.469)}\\ .428 / .329 \textit{(.280)}\\ .502 / .370 \textit{(.224)}\end{tabular}} & \multicolumn{1}{c|}{\begin{tabular}[c]{@{}c@{}}.201 / -.061 \textit{(-.090)}\\ .193 / .090 \textit{(.023)}\\ \textbf{.235} / .022 \textit{(-.076)}\end{tabular}} & \multicolumn{1}{c|}{\begin{tabular}[c]{@{}c@{}}\textbf{.499} / .295 \textit{(.225)}\\ .484 / .259 \textit{(.067)}\\ .307 / .170 \textit{(.224)}\end{tabular}} & \begin{tabular}[c]{@{}c@{}}.292 / .149 \textit{(.069)}\\ \textbf{.209} / .123 \textit{(.017)}\\ .162 / .012 \textit{(-.068)}\end{tabular} & \begin{tabular}[c]{@{}c@{}}\textbf{.418} / .216 \textit{(.279)}\\ .316 / .175 \textit{(.086)}\\ .378 / .247 \textit{(.209)}\end{tabular} & \multicolumn{1}{c|}{\begin{tabular}[c]{@{}c@{}}\textbf{.386} / .207 \textit{(.094)}\\ .247 / .058 \textit{(-.116)}\\ .358 / .224 \textit{(.130)}\end{tabular}} & \begin{tabular}[c]{@{}c@{}}\textbf{.329} / .028 \textit{(.314)}\\ .194 / -.105 \textit{(.035)}\\ .322 / .132 \textit{(-.100)}\end{tabular} & \multicolumn{1}{c|}{\begin{tabular}[c]{@{}c@{}}.466 / .227 \textit{(.011)}\\ \textbf{.539} / .275 \textit{(-.090)}\\ .465 / .035 \textit{(.030)}\end{tabular}} & \multicolumn{1}{c|}{\begin{tabular}[c]{@{}c@{}}\textbf{.671} / .587 \textit{(.165)}\\ .645 / .256 \textit{(465)}\\ .583 / .135 \textit{(.448)}\end{tabular}} \\ \cline{2-14} 
\multicolumn{1}{c|}{} & \multicolumn{1}{c|}{WiDiD} & \multicolumn{1}{c|}{\begin{tabular}[c]{@{}c@{}}BERT\\ mBERT\\ XLM-R\end{tabular}} & \multicolumn{1}{c|}{\begin{tabular}[c]{@{}c@{}}.635 / .441 \textit{(.385)}\\ .600 / .317 \textit{(.323)}\\ \textbf{.760} / .663 \textit{(.564)}\end{tabular}} & \multicolumn{1}{c|}{\begin{tabular}[c]{@{}c@{}}/\\ .252 / .055 \textit{(-.039)}\\ \textbf{.347} / -.077 \textit{(-.064)}\end{tabular}} & \multicolumn{1}{c|}{\begin{tabular}[c]{@{}c@{}}.465 / .322 \textit{(.355)}\\ .610 / .422 \textit{(.312)}\\ \textbf{.721} / .557 \textit{(.499)}\end{tabular}} & \multicolumn{1}{c|}{\begin{tabular}[c]{@{}c@{}}.432 / .177 \textit{(.106)}\\ \textbf{.521} / .413 \textit{(.195)}\\ .503 / .220 \textit{(.129)}\end{tabular}} & \multicolumn{1}{c|}{\begin{tabular}[c]{@{}c@{}}.466 / .361 \textit{(.383)}\\ \textbf{.575} / .272 \textit{(.343)}\\ .526 / .437 \textit{(.459)}\end{tabular}} & \begin{tabular}[c]{@{}c@{}}.388 / .136 \textit{(.135)}\\ .255 / .215 \textit{(-.068)}\\ \textbf{.426} / .223 \textit{(.268)}\end{tabular} & \begin{tabular}[c]{@{}c@{}}.410 / .190 \textit{(.102)}\\  .373 / .056 \textit{(.160)}\\ \textbf{.460} / .352 \textit{(.216)}\end{tabular} & \multicolumn{1}{c|}{\begin{tabular}[c]{@{}c@{}}.408 / .280 \textit{(.243)}\\ .327 / .252 \textit{(.142)}\\ \textbf{.485} / .304 \textit{(.342)}\end{tabular}} & \begin{tabular}[c]{@{}c@{}}\textbf{.531} / .160 \textit{(.233)}\\ .500 / .459 \textit{(.241)}\\ .505 / .399 \textit{(.226)}\end{tabular} & \multicolumn{1}{c|}{\begin{tabular}[c]{@{}c@{}}\textbf{.578} / .336 \textit{(.087)}\\ .467 / .292 \textit{(.290)}\\ .440 / .336 \textit{(.349)}\end{tabular}} & \multicolumn{1}{c|}{\begin{tabular}[c]{@{}c@{}}\textbf{.701} / .537 \textit{(.533)}\\ .620 / .513 \textit{(.338)}\\ .637 / .349 \textit{(.382)}\end{tabular}} \\ \cline{2-14} 
\end{tabular}%
}
\caption{\textbf{Top score for GCD} obtained using BERT, mBERT, and XLM-R. We present results for the optimal combination and the outcome obtained by summing the last four layers, separated by a slash (i.e., best results \textbf{/} sum of last four layers). Additionally, for comparison purposes, we include the result obtained using the last layer individually \textit{(enclosed in brackets).}. Top scores for approach and benchmark are highlighted in \textbf{bold}.}
\label{tab:best-vs-last4}
\end{table}
\end{landscape}

\end{document}